\documentclass[10pt,twocolumn,letterpaper]{article}

\usepackage[pagenumbers]{cvpr} 

\usepackage{graphicx}
\usepackage{amsmath}
\usepackage{amssymb}
\usepackage{booktabs}
\usepackage{xcolor}
\usepackage{multirow}
\usepackage{makecell}
\usepackage{wrapfig}
\usepackage{colortbl}
\usepackage{tikz}
\usepackage{twemojis}
\usepackage{fontawesome5}

\definecolor{darkgreen}{HTML}{11823b}
\definecolor{green}{HTML}{48bf53}
\definecolor{lightgreen}{HTML}{91f086}

\definecolor{lightred}{HTML}{F6BDC0}

\definecolor{gold}{HTML}{FBF2D2}
\definecolor{silver}{HTML}{DDDDDD}
\definecolor{bronze}{HTML}{EED2B8}

\definecolor{goldD}{HTML}{D9AE13}
\definecolor{silverD}{HTML}{909090}
\definecolor{bronzeD}{HTML}{9A5F26}

\definecolor{MS}{HTML}{b85450}
\definecolor{optical}{HTML}{6c8ebf}
\definecolor{student}{HTML}{1c8d39}
\definecolor{augs}{HTML}{9673a6}
\definecolor{text}{HTML}{d6b656}

\definecolor{falseNeg}{HTML}{1E88E5}
\definecolor{falsePos}{HTML}{E4AD06}

\newcommand{\medal}[3]{\tikz[baseline=(char.base)]{\node[rounded corners=2pt,fill=#1,draw=#2,inner sep=1.5pt](char){#3};}}

\newcommand{\bm}[2]{
    \ifcase#1\or
      {\medal{gold}{goldD}{\textbf{#2}}}
    \or 
      {\medal{silver}{silverD}{#2}}
    \or 
      {\medal{bronze}{bronzeD}{#2}}
    \else 
      #2
    \fi\ignorespaces
}

\definecolor{cvprblue}{rgb}{0.21,0.49,0.74}
\usepackage[pagebackref,breaklinks,colorlinks,allcolors=cvprblue]{hyperref}

\def\confName{CVPR}
\def\confYear{2027}

\title{Towards True Multispectral Vision-Language Models for Earth Observation}

\author{Filip Wolf
\and Blaž Rolih
\and Luka Čehovin Zajc
\and University of Ljubljana, Faculty of Computer and Information Science, Slovenia\\
{\tt\small filip.wolf@fri.uni-lj.si}
}

\begin{document}
\maketitle
\begin{abstract}
Vision-language models (VLMs) are highly attractive for Earth Observation (EO), where labeled data are scarce and tasks are diverse, as they allow downstream tasks to be specified in natural language. However, general-purpose VLMs such as \texttt{dino.txt} operate only on RGB inputs, ignoring the rich spectral information captured by EO sensors. Moreover, we show that existing multispectral (MS) EO VLMs perform no better with MS than with RGB inputs alone. To address this, we propose \textbf{LEO}, a multispectral EO VLM. We first train an MS visual backbone aligned with DINOv3 during pretraining, so that its MS features can be interpreted by the frozen \texttt{dino.txt} text encoder without any image-text alignment. We then adapt only the vision-side class projection heads on MS image-text pairs, improving zero-shot classification while preserving dense capabilities. Optionally, a patch loss derived from land-cover maps adds dense guidance, yielding LEO$_\text{LC}$. On seven MS open-vocabulary semantic segmentation and zero-shot classification datasets, LEO outperforms the best competing method trained without auxiliary semantic guidance by 7.8 points and DOFA-CLIP, the best competing method trained with it, by 3.1 points, while gaining 5.2 points from MS over RGB inputs. LEO$_\text{LC}$ further extends this lead. 
\end{abstract}    
\section{Introduction}
\label{sec:intro}

Earth Observation Foundation Models (EOFMs) provide strong visual representations for a wide range of tasks~\cite{wolf2026deo,herzog2026olmoearth,pmlr-v267-tseng25a,jakubik2025terramind}. However, their use typically requires task-specific prediction heads trained on annotated data. Vision-Language Models (VLMs) instead offer a complementary paradigm by aligning visual representations with natural language, enabling tasks such as classification and segmentation to be specified via textual descriptions without requiring labeled examples for each downstream dataset~\cite{radford2021learning,zhai2023sigmoid,jose2025dinov2}. This flexibility is particularly attractive for Earth Observation (EO), where task requirements are diverse and evolving~\cite{simumba2025geo}. In addition, language-aligned visual representations naturally fit emerging geospatial agents that combine natural-language task specifications and planning with specialized analytical tools~\cite{shabbir2026openearthagent,zhou2026geopilot}.

\begin{figure}[t]
  \centering
  \includegraphics[width=\linewidth]{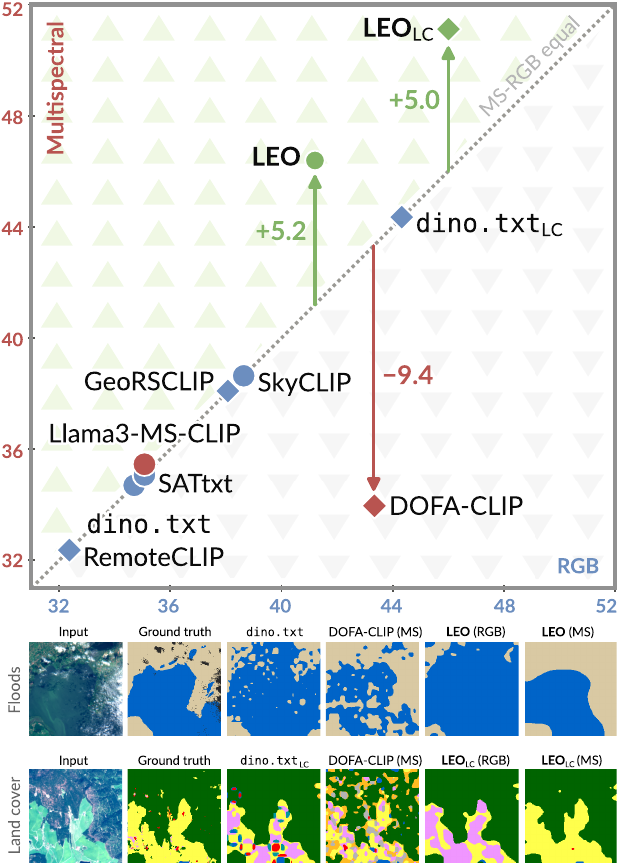}
  \caption{\textbf{Existing multispectral VLMs do not benefit from multispectral information.} We show average OVSS and ZSC performance across seven MS datasets with full MS inputs and RGB-only inputs. \textcolor{MS}{Red} methods support MS inputs, while \textcolor{optical}{blue} methods are RGB-only. Circles represent methods trained without auxiliary semantic guidance, while diamonds represent methods trained with it. \textbf{LEO} is the only method to benefit from additional MS information, sitting above the diagonal.  At the bottom, we show qualitative examples of floods and land cover OVSS.}
  \label{fig:first}
\end{figure}

However, most general-purpose VLMs operate on RGB imagery, whereas EO sensors capture additional spectral information valuable for environmental monitoring and land-cover analysis. Recent EO VLMs have begun incorporating multispectral (MS) observations~\cite{marimo2025beyond,xiong2025dofa,do2026spectrally}, yet the benefits of additional spectral bands remain inconsistent. As we demonstrate empirically, several existing methods achieve comparable or even better performance when restricted to RGB inputs (\Cref{fig:first}). \textit{Supporting multispectral inputs, therefore, does not necessarily imply effective use of spectral information in language-aligned representations.}

Modern EOFMs, such as DEO~\cite{wolf2026deo}, demonstrate the benefits of incorporating knowledge from general-purpose VFMs, such as DINOv3~\cite{simeoni2025dinov3}, into multispectral representation learning. In a similar fashion, text-aligned counterparts of VFMs, \eg, \texttt{dino.txt}~\cite{jose2025dinov2}\footnote{We will use \texttt{dino.txt} to refer to the text encoder, and DINOv3 to refer to the visual backbone.}, could bootstrap EO VLMs, as aligning an encoder for a new modality with the representation space of a frozen VLM has been shown to transfer its zero-shot capabilities~\cite{girdhar2023imagebind,zhu2024languagebind}. However, this requires careful alignment: visual representations must reside in the same space the text encoder was aligned with~\cite{girdhar2023imagebind,jose2025dinov2}, while subsequent image-text alignment must not overwrite pretrained visual and textual representations~\cite{zhai2022lit}. As our experiments show, distilling an RGB VFM into a multispectral backbone using only RGB inputs renders its MS features uninterpretable by the text encoder, and subsequent image-text alignment cannot fully compensate for this.

To address these limitations, we propose \textbf{LEO} (\textit{\textbf{L}anguage for \textbf{E}arth \textbf{O}bservation}). We start by creating a DINOv3-aligned multispectral visual backbone by contrastively training a unified MS-RGB representation space and simultaneously aligning it with DINOv3. This enables \texttt{dino.txt} to interpret MS features while retaining RGB capabilities without any image-text alignment. We then adapt only the vision-side class projection heads on MS image-text pairs, keeping both encoders and the patch heads frozen, as we find that global image-text alignment otherwise degrades dense capabilities. Additionally, we introduce an optional patch loss derived from land-cover maps that provides dense guidance to the vision-side patch projection heads, yielding LEO$\mathrm{_{LC}}$. On seven diverse MS zero-shot classification (ZSC) and open-vocabulary semantic segmentation (OVSS) datasets, LEO outperforms the best competing method trained without auxiliary semantic guidance by 7.8 points and DOFA-CLIP~\cite{xiong2025dofa}, the best competing method trained with it, by 3.1 points. Notably, it also surpasses \texttt{dino.txt} trained with the same image-text alignment by an average of 10.4 points, demonstrating the advantage of our pretrained visual backbone. Importantly, LEO performs, on average, 5.2 points better with MS inputs than with RGB alone, making it the only method we evaluated that benefits from additional MS bands. LEO$\mathrm{_{LC}}$ further outperforms DOFA-CLIP by 7.7 points, and \texttt{dino.txt} trained with the same land-cover dense guidance by 6.6 points.

\begin{figure*}[t]
  \centering
  \includegraphics[width=\textwidth]{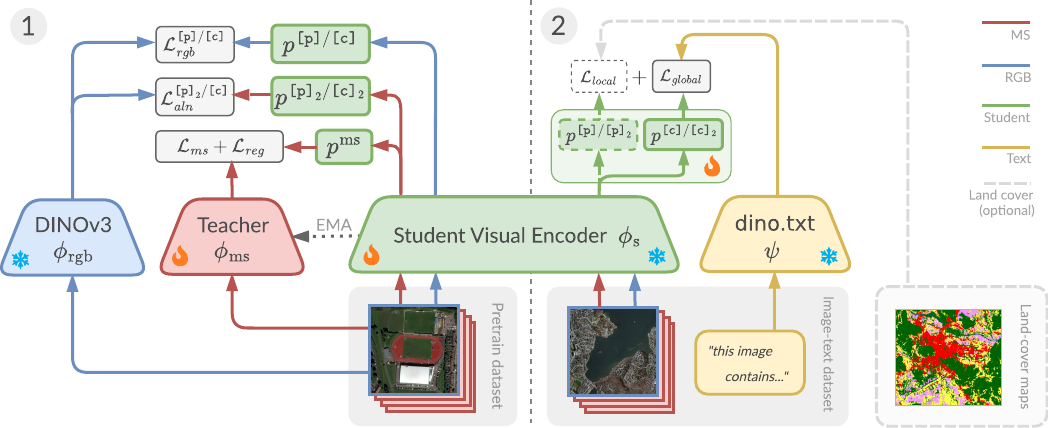}
  \caption{\textbf{LEO method overview.} On the left side of the image in \textbf{Stage 1}, a student (\textcolor{student}{green}) is aligned with a multispectral teacher (\textcolor{MS}{red}) to teach MS representations and with DINOv3~\cite{simeoni2025dinov3} (\textcolor{optical}{blue}) with both RGB and MS features to align across both spaces, creating the \textbf{LEO(v)} visual backbone. On the right side in \textbf{Stage 2}, the same student is aligned on MS image-caption pairs using \texttt{dino.txt}~\cite{jose2025dinov2} (\textcolor{text}{yellow}) by training its class projection heads, thereby creating \textbf{LEO}. Optionally, a land cover-derived patch loss is used to increase dense performance, creating LEO$\mathrm{_{LC}}$.}
  \label{fig:method_diagram}
\end{figure*}

\noindent Overall, our contributions are as follows:
\begin{itemize}
    \item We show that existing MS-capable EO VLMs fail to benefit from MS bands (\Cref{fig:first}), tracing this failure to their visual backbones and image-text alignment (\Cref{subsec:linear_probing,subsec:ms_vs_rgb}). 
    \item We demonstrate that carefully aligning a multispectral visual backbone with the visual backbone of a pretrained VLM during pretraining \textit{enables the VLM's text encoder to interpret MS features without any image-text alignment} (\Cref{subsec:ms_vision_backbone}).
    \item Building on these insights, we train \textbf{LEO}, a state-of-the-art multispectral EO VLM that effectively leverages MS information, using our proposed two-stage training scheme (\Cref{fig:method_diagram}), and validate it through extensive comparisons and ablations.
\end{itemize}

\section{Related Work}
\label{sec:rl}

\noindent\textbf{Vision-Language Models.} VLMs learn a joint embedding space for images and text by aligning visual and textual encoders on image-text datasets~\cite{radford2021learning,jia2021scaling,zhai2023sigmoid}. While CLIP-style models perform well on zero-shot classification (ZSC), they perform poorly on dense tasks such as open-vocabulary semantic segmentation (OVSS), motivating methods that improve patch-level alignment~\cite{zhou2022extract,zhong2022regionclip,barsellotti2025talking,maninis2025tips,cao2026tipsv2,tschannen2025siglip}. Among these, \texttt{dino.txt}~\cite{jose2025dinov2} aligns a text encoder with a frozen DINO backbone~\cite{oquab2024dinov,simeoni2025dinov3} and achieves strong global and dense capabilities.

\noindent\textbf{Binding new modalities.} ImageBind~\cite{girdhar2023imagebind} shows that a new modality can inherit zero-shot capabilities without text supervision by aligning its encoder with the image space of a frozen VLM, while LanguageBind~\cite{zhu2024languagebind} aligns each modality directly with a frozen text encoder. Both train a full encoder for each modality on large-scale paired data. Instead, we bind MS to the image space of \texttt{dino.txt} during pretraining on a single backbone shared with RGB using unlabeled imagery. Consequently, the subsequent image-text alignment is lightweight: both encoders remain frozen, and only the vision-side projection heads are trained on a relatively small set of MS image-text pairs.

\noindent\textbf{Earth Observation VLMs.} SkyCLIP~\cite{wang2024skyscript}, RemoteCLIP~\cite{liu2024remoteclip}, and GeoRSCLIP~\cite{zhang2024rs5m} adapt CLIP to RGB EO imagery paired with short captions or category names. Llama3-MS-CLIP~\cite{marimo2025beyond} and DOFA-CLIP~\cite{xiong2025dofa} introduced multispectral support; however, as our experiments show, they fail to leverage the additional bands and perform no better with MS than with RGB inputs alone. SATtxt~\cite{do2026spectrally} trains projectors on top of a DINOv3 backbone and pairs it with a large LLM as the text encoder, but does not use MS inputs at inference. In contrast, LEO operates directly on MS inputs and requires no LLM, as it retains the compact \texttt{dino.txt} text encoder, which is already aligned with DINOv3. It also supports OVSS, which remains largely unexplored for multispectral EO VLMs.
\section{LEO}

We aim to create a multispectral Earth Observation Vision-Language Model (EO VLM) by building on \texttt{dino.txt}~\cite{jose2025dinov2}, the text-aligned counterpart of DINOv3~\cite{simeoni2025dinov3}. We therefore: \textbf{(1)} pretrain an MS visual backbone aligned with DINOv3, so that \texttt{dino.txt} can interpret its MS features directly, and \textbf{(2)} keep both encoders frozen and adapt only the vision-side class projection heads on MS image-text pairs. Optionally, a patch loss derived from land-cover maps provides additional dense guidance. A diagram of our method is shown in \Cref{fig:method_diagram}.

\subsection{Preliminaries}
\label{subsec:prelim}

\noindent\textbf{\texttt{dino.txt}.} To leverage the strong visual capabilities of DINOv2~\cite{oquab2024dinov} in zero-shot and open-vocabulary settings, the authors of \texttt{dino.txt}~\cite{jose2025dinov2} align it with a textual encoder on a large, web-crawled, curated image-text dataset. To prevent catastrophic forgetting of visual concepts, only the textual side is trained with CLIP loss~\cite{radford2021learning} in a Locked-image text Tuning (LiT)~\cite{zhai2022lit} fashion, alongside a two-layer transformer~\cite{vaswani2017attention} vision head on the visual side, which we use in all \texttt{dino.txt} evaluations except linear probing. Crucially, to preserve the remarkable dense capabilities of DINOv2, patch tokens are pooled and, together with the class token, aligned with the textual features. The resulting VLM achieves strong results across a broad range of downstream tasks. Later, a DINOv3-aligned variant is released~\cite{simeoni2025dinov3}, which we refer to in this paper.

\noindent\textbf{DEO.} To train an EOFM with effective MS capabilities while leveraging the strong RGB representations of DINOv3, the authors of DEO~\cite{wolf2026deo} align a student $\phi_\mathrm{s}$ to both an MS teacher $\phi_\mathrm{ms}$ and an RGB teacher $\phi_\mathrm{rgb}$, specifically DINOv3. An MS image $\mathrm{\textbf{I}_{ms}}\in\mathbb{R}^{C\times H \times W}$ or its subset RGB image $\mathrm{\textbf{I}_{rgb}}\in\mathbb{R}^{3\times H \times W}$, with $C$ the number of spectral channels, is first passed to $\phi_\mathrm{s}$ to obtain the corresponding student representations $\phi_\mathrm{s}(\textbf{I}_\mathrm{rgb}) = \mathrm{\textbf{z}_s^{rgb}}$ and $\phi_\mathrm{s}(\textbf{I}_\mathrm{ms}) = \mathrm{\textbf{z}_s^{ms}}$. The MS teacher $\phi_\mathrm{ms}$ receives $\mathrm{\textbf{I}_{ms}}\in\mathbb{R}^{C\times H \times W}$ to obtain teacher MS representations $\phi_\mathrm{ms}(\textbf{I}_\mathrm{ms}) = \mathrm{\textbf{z}_t^{ms}}$, and DINOv3 receives $\mathrm{\textbf{I}_{rgb}}\in\mathbb{R}^{3\times H \times W}$ to obtain patch and \texttt{[cls]} token features $\phi_\mathrm{rgb}(\textbf{I}_\mathrm{rgb}) = \mathrm{[\textbf{z}_{d}^{\texttt{[p]}}, \textbf{z}_{d}^{\texttt{[c]}}]}$ (without the \texttt{dino.txt} vision head). The student and MS teacher representations are then passed through trainable projection heads $p$, implemented as multi-layer MLPs (details in the Supplementary), to align their feature dimensions and help prevent collapse~\cite{caron2021emerging}. MS capabilities are trained in a self-supervised manner by aligning projected MS features $p_\mathrm{s}^\mathrm{ms}(\mathrm{\textbf{z}_s^{ms}})$ and $p_\mathrm{t}^\mathrm{ms}(\mathrm{\textbf{z}_t^{ms}})$, and RGB capabilities by aligning projected RGB patch and \texttt{[cls]} token features $p_\mathrm{s}^{\texttt{[p]}}(\mathrm{\textbf{z}_s^{rgb}}[\texttt{p}])$ and $\mathrm{\textbf{z}_{d}^{\texttt{[p]}}}$, and $p_\mathrm{s}^{\texttt{[c]}}(\mathrm{\textbf{z}_s^{rgb}}[\texttt{c}])$ and $\mathrm{\textbf{z}_{d}^{\texttt{[c]}}}$ via a cosine loss. A regularization term on the student MS features $\mathrm{\textbf{z}_s^{ms}}$ is used to prevent collapse. Concretely:
\begin{equation}
\begin{aligned}
\mathcal{L_\mathrm{DEO}} 
 &= \mathcal{L}_{ms}\big(p_\mathrm{s}^\mathrm{ms}(\mathrm{\textbf{z}_s^{ms}}), p_\mathrm{t}^\mathrm{ms}(\mathrm{\textbf{z}_t^{ms}})\big) \\
& + \mathcal{L}_{rgb}^{\texttt{[p]}}\big(p_\mathrm{s}^{\texttt{[p]}}(\mathrm{\textbf{z}_s^{rgb}}[\texttt{p}]), \mathrm{\textbf{z}_d^{[p]}}\big) \\
& + \mathcal{L}_{rgb}^{\texttt{[c]}}\big(p_\mathrm{s}^{\texttt{[c]}}(\mathrm{\textbf{z}_s^{rgb}}[\texttt{c}]), \mathrm{\textbf{z}_d^{[c]}}\big) + \mathcal{L}_{reg}\big(\mathrm{\textbf{z}_s^{ms}}\big).
\end{aligned}
\label{eq:deo}
\end{equation}
Following a contrastive self-distillation training approach, EMA is used to update the $\phi_\mathrm{ms}$ weights from $\phi_\mathrm{s}$~\cite{wu2025simplifying,caron2021emerging}. $\phi_\mathrm{rgb}$ is kept frozen because its primary functions are DINOv3 alignment and distilling RGB capabilities. 

\begin{table}[t]
\centering
\begin{tabular}{lcccc}
\toprule
 & \multicolumn{2}{c}{\textbf{MS}} & \multicolumn{2}{c}{\textbf{RGB}} \\
\cmidrule(lr){2-3} \cmidrule(lr){4-5}
Vision backbone & OVSS & ZSC & OVSS & ZSC \\
\midrule
DINOv3~\cite{simeoni2025dinov3,jose2025dinov2} & - & - & \textcolor{darkgreen}{36.6} & \textcolor{darkgreen}{32.1} \\
DEO~\cite{wolf2026deo} & \textcolor{red}{20.8} & \textcolor{red}{10.4} & \textcolor{darkgreen}{41.3} & \textcolor{darkgreen}{28.9} \\
LEO(v) & \textcolor{darkgreen}{\textbf{46.4}} & \textcolor{darkgreen}{\textbf{35.2}} & \textcolor{darkgreen}{40.5} & \textcolor{darkgreen}{29.2} \\
\bottomrule
\end{tabular}
\caption{We evaluate visual backbones using \texttt{dino.txt}~\cite{jose2025dinov2} on seven MS datasets, using both full MS bands and only RGB inputs, and report average OVSS and ZSC results. Note that DINOv3 accepts only RGB as input and is evaluated with its \texttt{dino.txt} vision head~\cite{jose2025dinov2}, while DEO and LEO(v) use their projection heads. We retrain DEO~\cite{wolf2026deo}; results for the original DEO and the full table are in the Supplementary.}
\label{tab:ov_dinov3_alignment}
\end{table}

\subsection{Multispectral Vision Backbone}
\label{subsec:ms_vision_backbone}

Multispectrality is a prominent aspect of EO imagery and can be naturally encoded in the visual backbone of a VLM. However, for \texttt{dino.txt} to read its outputs, the visual backbone must operate in the same feature space as the visual backbone \texttt{dino.txt} was originally aligned with, namely DINOv3 (\Cref{subsec:prelim}). These two requirements can be met either by adapting DINOv3 to the MS domain or by aligning an MS-capable visual backbone with DINOv3. We opt for the latter, as aligning an existing backbone with DINOv3 (typically via distillation) is an effective, proven approach~\cite{wolf2026deo,wang2025towards}. For this purpose, we adopt the recent EOFM training scheme used in DEO~\cite{wolf2026deo} (\Cref{subsec:prelim}), as its distillation-oriented pretraining enables effective alignment with DINOv3 while providing strong MS performance.

\begin{figure}[t]
  \centering
  \includegraphics[width=\linewidth]{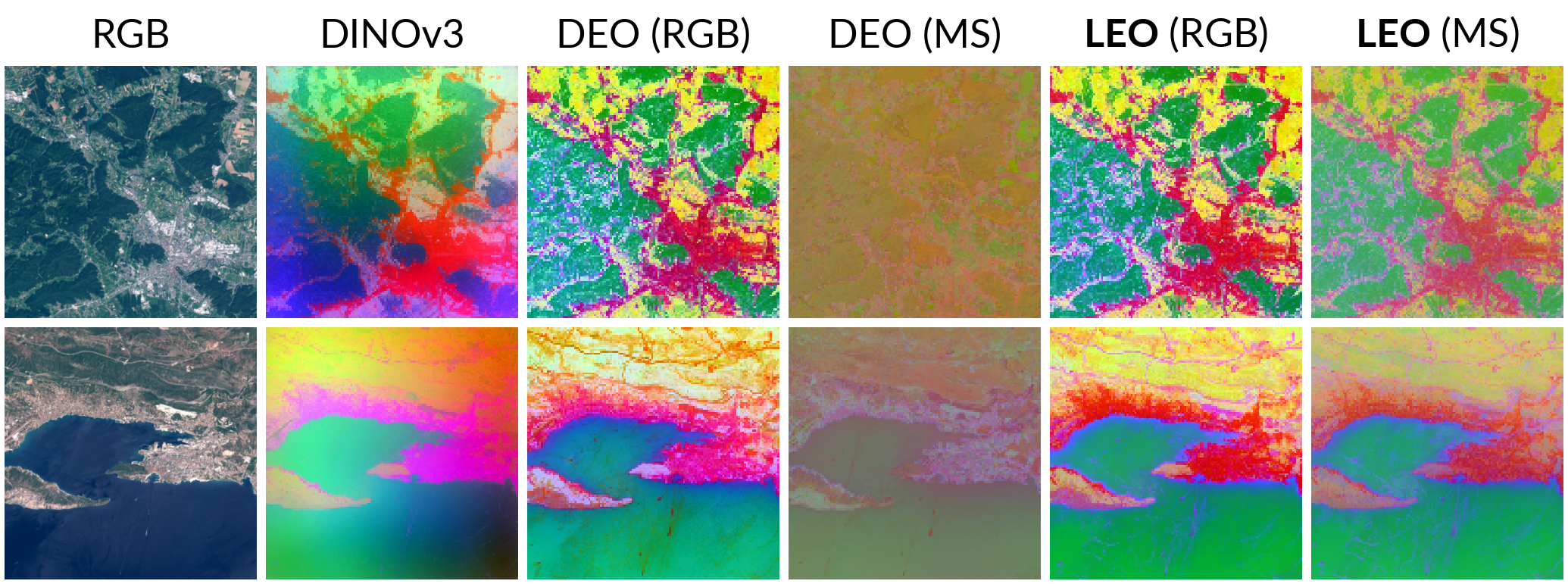}
  \caption{\textbf{Visual backbone feature PCA}. PCA is fitted separately for each model. LEO’s RGB and MS features resemble DINOv3’s spatial structure, while DEO’s MS features do not.}
  \label{fig:pca}
\end{figure}

We retrain DEO with a DINOv3-initialized ViT-Base~\cite{dosovitskiy2020image} backbone. To gauge its alignment with DINOv3, and by extension \texttt{dino.txt}, we evaluate it in zero-shot and open-vocabulary settings on seven semantic segmentation and classification datasets against DINOv3 in \Cref{tab:ov_dinov3_alignment} (see \Cref{sec:results} for more evaluation details). Each dataset is evaluated with both its full MS input and its RGB bands only, and we pass our backbone features through the DINOv3-aligned projection heads $p_\mathrm{s}^{\texttt{[p]}}$ and $p_\mathrm{s}^{\texttt{[c]}}$, while for the DINOv3 baseline, we instead use the \texttt{dino.txt} vision head~\cite{jose2025dinov2}. We see that this setup recovers the DINOv3 baseline performance on RGB inputs (row 2); however, the results collapse for MS inputs. We attribute this to the way DEO is trained: DINOv3 is aligned only with projected student RGB features $p_\mathrm{s}^{\texttt{[p]}/\texttt{[c]}}(\mathrm{\textbf{z}_s^{rgb}})$ (terms 2 and 3 in \Cref{eq:deo}). We therefore modify DEO by additionally aligning MS student features $\mathrm{\textbf{z}_s^{ms}}$ and new projection heads $p_\mathrm{s}^{\texttt{[p]}_2/\texttt{[c]}_2}$ with DINOv3 features $\mathrm{\textbf{z}_{d}}$ using cosine losses $\mathcal{L}_{aln}^{\texttt{[p]}}$ and $\mathcal{L}_{aln}^{\texttt{[c]}}$, indicated in green:
\begin{equation}
\begin{aligned}
 \mathcal{L_\mathrm{LEO(v)}}& = \mathcal{L_\mathrm{DEO}}
+ \textcolor{darkgreen}{\mathcal{L}_{aln}^{\texttt{[p]}}\big(p_\mathrm{s}^{\texttt{[p]}_2}(\mathrm{\textbf{z}_s^{ms}}[\texttt{p}]), \mathrm{\textbf{z}_d^{[p]}}\big)} \\
 &+ \textcolor{darkgreen}{\mathcal{L}_{aln}^{\texttt{[c]}}\big(p_\mathrm{s}^{\texttt{[c]}_2}(\mathrm{\textbf{z}_s^{ms}}[\texttt{c}]),\mathrm{\textbf{z}_d^{[c]}}\big)}.
\end{aligned}
\end{equation}
where we refer to this new visual backbone as \textbf{LEO(v)}.
We show this process in Stage 1 of \Cref{fig:method_diagram} and visualize it in \Cref{fig:pca}: with $\mathcal{L}_{aln}$, the MS features of LEO(v) share the structure of its RGB features, whereas DEO’s MS features do not. Training details are in \Cref{subsec:training_details}.
As shown in row 3 of \Cref{tab:ov_dinov3_alignment}, when paired with \texttt{dino.txt}, \textit{LEO(v) maintains RGB performance while greatly improving MS performance, which is now significantly above RGB performance, all without any dedicated image-text alignment}.

\subsection{Multispectral Image-Text Alignment}
\label{subsec:lit}

However, as we show in \Cref{tab:ablation}, \texttt{dino.txt} does not yet fully exploit the MS representations of LEO(v), especially for ZSC. We therefore adapt the vision-side projection heads using a global CLIP~\cite{radford2021learning} loss on MS image-text pairs. First, we use the vision encoder $\phi$ with the MS projection heads $p_\mathrm{s}^{\texttt{[p]}_2/\texttt{[c]}_2}$ to obtain features from an MS input image $\mathrm{\textbf{I}}\in\mathbb{R}^{C\times H \times W}$.  This yields $p_\mathrm{s}^{\texttt{[c]}_2/\texttt{[p]}_2}\big(\phi(\mathrm{\textbf{I}})\big) = [\textbf{c}, \textbf{f}_1, ..., \textbf{f}_N]$, where $\textbf{c}$ is the \texttt{[cls]} token and $\textbf{f}_1, ..., \textbf{f}_N$ are the patch tokens of dimension $D$. 
For the global CLIP loss, as in \texttt{dino.txt}, the patch tokens are pooled via a pooling function $\mu$ and concatenated along the feature dimension with the \texttt{[cls]} token to form $\mathrm{\textbf{v}}=[\textbf{c}; \mu (\textbf{f}_1, ..., \textbf{f}_N)]$, where $[\cdot;\cdot]$ denotes feature concatenation, yielding a feature vector of dimension $2D$.

We then use the textual encoder $\psi$ to obtain representations from a caption: $\psi(\texttt{text}) = \mathrm{\textbf{t}}$. Due to the way \texttt{dino.txt} was trained (\Cref{subsec:prelim}), $\mathrm{\textbf{t}}$ consists of a patch-aligned part $\mathrm{\textbf{t}} [D:2D]$ and a \texttt{[cls]}-aligned part $\mathrm{\textbf{t}} [:D]$, matching the dimension of the image embedding $\mathrm{\textbf{v}}$. Finally, we compute the symmetric CLIP loss~\cite{radford2021learning} over a batch of $B$ image-text pairs:
\begin{equation}
  \mathcal{L}_{\mathrm{global}} = -\sum_{i}^{B} \Big( \log \sigma(\textbf{s}_{i,:})_i + \log \sigma(\textbf{s}_{:,i})_i \Big),
  \label{eq:global}
\end{equation}
where $\textbf{s}_{i,j} = \cos(\textbf{v}_i, \textbf{t}_j) / \tau$ is the temperature-scaled similarity between image $i$ and caption $j$, and $\sigma(\cdot)_i$ denotes the $i$-th entry of the softmax. The two terms normalize over captions and over images, respectively. Our experiments in \Cref{tab:ablation} show that training the patch heads $p_\mathrm{s}^{\texttt{[p]}_2}$ degrades dense performance; therefore, we train only the class projection heads $p_\mathrm{s}^{\texttt{[c]}_2}$. This yields our multispectral VLM \textbf{LEO}, shown in Stage 2 of \Cref{fig:method_diagram}.

\subsection{Local Concept Alignment}
\label{subsec:lca}

The above recipe creates a strong MS VLM that performs well on open-vocabulary and zero-shot tasks. However, following related methods~\cite{xiong2025dofa, liu2024remoteclip}, dense performance can be further improved with guidance from freely available land-cover maps. We use ESA's WorldCover~\cite{zanaga2022worldcover}, which assigns each pixel one of 11 land-cover labels and is spatially aligned with our MS training data (\Cref{subsec:training_details}), as \textit{weak} dense guidance that encourages patches of the same land-cover class to share similar representations.

We first map WorldCover pixels to our visual backbone patches. A patch is retained if at least half of its pixels are labeled and its majority class covers at least half of them, in which case it is assigned that class, while all other patches are discarded. We then pull patches of the same class together and push patches of different classes apart with a supervised contrastive loss~\cite{khosla2020supervised,wang2021exploring}:
\begin{equation}
  \mathcal{L}_{\mathrm{local}} = - \sum_{a \in A} \sum_{p \in P(a)} \log \sigma\big(\cos(\mathbf{f}_a, \mathbf{f}_k) / \tau\big)_p,
\end{equation}
where $A$ is a set of patches from the whole batch,  $P(a) \subset A \setminus \{a\}$ denotes the patches with the same class as patch $a$, and $\sigma(\cdot)_p$ is the softmax over all other patches $k \in A \setminus \{a\}$, evaluated at patch $p$. Unlike $\mathcal{L}_\mathrm{global}$, this loss operates only on image patches, updates only the patch projection heads, and does not rely on text. Finally, we jointly minimize the two losses:
\begin{equation}
  \mathcal{L} = \mathcal{L}_{\mathrm{global}} + \mathcal{L}_{\mathrm{local}},
\end{equation}
yielding the land-cover-enhanced variant of our model: \textbf{LEO$_\mathrm{LC}$}.

\subsection{Training Details}
\label{subsec:training_details}

LEO(v) pretraining follows DEO~\cite{wolf2026deo} using a mix of images from the fMoW-Sentinel~\cite{cong2022satmae} and fMoW-RGB~\cite{christie2018functional} datasets for 100 epochs with the Adam optimizer on 4 B300 GPUs, with a total batch size of 256. For Sentinel-2 imagery, we remove the three 60m atmospheric bands, leaving 10 bands.
For image-text alignment, we use the SSL4EO-S12 dataset, which comprises 976k Sentinel-2 images, paired with Llama3-generated captions~\cite{wang2023ssl4eo,marimo2025beyond}, and train for 1 epoch at a learning rate of $5 \times 10^{-5}$, a batch size of 256, and $\tau$ of 0.2.  We train 25\% of the time on RGB images to improve robustness on RGB datasets; in this case, we train $p_\mathrm{s}^\texttt{[c]}$ instead of $p_\mathrm{s}^{\texttt{[c]}_2}$. We apply cropping and flipping as image augmentations.
\newcommand{\g}[1]{\textcolor{gray}{#1}}
\newcommand{\tup}[1]{\tiny\textcolor{darkgray}{$\uparrow$#1}}
\definecolor{oursblue}{HTML}{E6EDF7}
\begin{table*}[t]
\centering
\setlength{\tabcolsep}{4pt}
\begin{tabular}{lllcc|cccccccc}
\toprule
& & & \multicolumn{2}{c}{\textbf{RGB}} & \multicolumn{7}{c}{\textbf{MS}} & \\
\cmidrule(lr){4-5} \cmidrule(lr){6-13}
& & & \multicolumn{2}{c}{OVSS} & \multicolumn{4}{c}{OVSS} & \multicolumn{3}{c}{Zero-shot classification} & \\
\cmidrule(lr){4-5} \cmidrule(lr){6-9} \cmidrule(lr){10-12}
Method & \smash{\makecell[b]{Vision\\Backbone}} & Input & Chesa. & pv4ger & S1F11 & WF & DFC & DEN & EuroSAT & BEN & So2Sat & \textit{Avg} \\
\midrule
\texttt{dino.txt}~\cite{jose2025dinov2} & \multirow{3}{*}{\small ViT-L/16} & \multirow{4}{*}{\small\textcolor{darkgray}{RGB}} & \bm2{30.6} & 67.7 & 52.6 & \bm2{56.3} & 24.1 & 13.4 & 56.9 & 25.2 & 14.3 & 34.7 \\
\texttt{dino.txt}$\mathrm{_{LEO}}$      & & & \bm1{32.5} & 62.8 & 44.4 & 47.7 & 28.1 & 13.8 & 70.1 & 23.7 & 24.1 & 36.0 \\
SATtxt~\cite{do2026spectrally}          & & & 28.9 & 19.9 & 39.4 & 43.1 & 28.1 & 11.3 & \bm2{75.6} & 26.5 & 21.4 & 35.1 \\
SkyCLIP~\cite{wang2024skyscript}        & \small ViT-L/14 & & 26.6 & \bm2{68.5} & \bm2{54.5} & 52.7 & 24.6 & 12.6 & 71.9 & \bm1{29.6} & \bm2{24.5} & \bm2{38.6} \\
\multirow{2}{*}{\shortstack[l]{Llama3-\\MS-CLIP~\cite{marimo2025beyond}}} & \multirow{2}{*}{\small ViT-B/16}
 & \small\g{RGB} & 27.8 & \bm1{69.2} & \g{50.0} & \g{51.8} & \g{27.9} & \g{15.5} & \g{60.4} & \g{23.5} & \g{16.3} & \g{35.1} \\
 & & \small\textcolor{darkgray}{MS}  & - & - & 50.7 & 52.6 & \bm1{29.4} & \bm1{16.3} & 57.4 & 23.8 & 17.9 & 35.5 \\
\midrule
\rowcolor{oursblue}
 & & \small\g{RGB} & 26.6 & 67.3 & \g{60.3} & \g{65.4} & \g{26.7} & \g{9.6} & \g{75.2} & \g{24.8} & \g{26.5} & \g{41.2} \\
\rowcolor{oursblue}
\multirow{-2}{*}{\textbf{LEO}} & \multirow{-2}{*}{\small ViT-B/16}
 & \small\textcolor{darkgray}{MS}  & - & - & \bm1{68.7} & \bm1{71.8} & \bm2{28.8} & \bm2{16.1} & \bm1{81.3} & \bm2{26.8} & \bm1{31.1} & \bm1{46.4} \\
\midrule
\midrule
\multicolumn{13}{l}{\textit{Auxiliary semantic guidance}} \\
\midrule
\texttt{dino.txt}$\mathrm{_{LC}}$       & \small ViT-L/16 & \multirow{3}{*}{\small\textcolor{darkgray}{RGB}} & \bm1{42.6} & 67.6 & \bm2{60.9} & \bm2{65.0} & \bm2{40.7} & \bm2{23.4} & 71.0 & 24.2 & \bm2{25.5} & \bm2{44.4} \\
GeoRSCLIP~\cite{zhang2024rs5m}          & \multirow{2}{*}{\small ViT-L/14} & & 32.1 & \bm2{70.0} & 50.7 & 54.2 & 28.7 & 15.1 & 69.7 & \bm1{30.6} & 17.6 & 38.1 \\
RemoteCLIP~\cite{liu2024remoteclip}     & & & 28.9 & 50.6 & 45.1 & 49.9 & 20.0 & 14.1 & 56.5 & 25.5 & 15.3 & 32.3 \\
\multirow{2}{*}{DOFA-CLIP~\cite{xiong2025dofa}} & \multirow{2}{*}{\small SO400M}
 & \small\textcolor{darkgray}{RGB} & 36.6 & \bm1{72.1} & 59.9 & 55.5 & 33.3 & 22.5 & \bm2{78.8} & \bm2{27.9} & 25.4 & 43.3 \\
 & & \small\g{MS}  & - & - & \g{58.3} & \g{48.6} & \g{20.4} & \g{13.3} & \g{58.5} & \g{23.3} & \g{14.9} & \g{33.9} \\
\rowcolor{oursblue}
 & & \small\g{RGB} & \bm2{40.1} & 61.3 & \g{61.1} & \g{65.6} & \g{39.8} & \g{25.4} & \g{78.2} & \g{25.3} & \g{26.7} & \g{46.0} \\
\rowcolor{oursblue}
\multirow{-2}{*}{\textbf{LEO}$\mathrm{_{LC}}$} & \multirow{-2}{*}{\small ViT-B/16}
 & \small\textcolor{darkgray}{MS}  & - & - & \bm1{71.5} & \bm1{74.0} & \bm1{43.3} & \bm1{26.6} & \bm1{81.2} & 27.3 & \bm1{33.3} & \bm1{51.0} \\
\bottomrule
\end{tabular}
\caption{\textbf{Open-vocabulary semantic segmentation and zero-shot classification results on seven MS and two RGB datasets.} Methods take either RGB or MS data as input and are divided into two groups: methods trained without auxiliary semantic guidance, and methods trained with it. \textit{Avg} covers MS datasets only. For S1F11~\cite{bonafilia2020sen1floods11}, WF~\cite{portales2023global}, DFC~\cite{9369830}, DEN~\cite{simumba2025geo,toker2022dynamicearthnet}, Chesa.~\cite{lacoste2023geo,robinson2019large}, and pv4ger~\cite{mayer20223d,lacoste2023geo}, we report mIoU; for EuroSAT~\cite{lacoste2023geo,helber2019eurosat} and So2Sat~\cite{zhu2020so2sat, lacoste2023geo}, we report Top-1 accuracy; and for BEN~\cite{lacoste2023geo,sumbul2019bigearthnet}, we report mAP. For MS-enabled methods Llama3-MS-CLIP~\cite{marimo2025beyond}, DOFA-CLIP~\cite{xiong2025dofa}, and LEO/LEO$\mathrm{_{LC}}$, we report both MS and RGB results, with secondary results in gray. \texttt{dino.txt}$\mathrm{_{LEO}}$ is \texttt{dino.txt}~\cite{jose2025dinov2} with its vision head trained using our image-text alignment, as it lacks separate class and patch heads, while \texttt{dino.txt}$\mathrm{_{LC}}$ follows the same protocol as LEO$\mathrm{_{LC}}$. For LEO/LEO$\mathrm{_{LC}}$ and both \texttt{dino.txt} runs, we show 3-run averages. \textcolor{goldD}{First} and \textcolor{silverD}{second} place results are marked. 3-run standard deviation, and dataset and method details are in the Supplementary.}
\label{tab:main}
\end{table*}

\section{Experiments}
\label{sec:results}

Our experiments address three questions: whether LEO outperforms existing EO VLMs on ZSC and OVSS (\Cref{subsec:main_results}), whether it benefits from MS inputs where existing methods do not (\Cref{subsec:ms_vs_rgb,subsec:linear_probing}), and which components of our method are responsible (\Cref{subsec:ablation}). We evaluate on 7 MS and 2 RGB datasets, following the protocol described in \Cref{subsec:evaluation_methodology}.

\subsection{Evaluation Methodology}
\label{subsec:evaluation_methodology}

All methods are evaluated at the resolution selected on the validation splits, using the 105 prompt templates from~\cite{zhang2024rs5m} with text embeddings averaged across templates.

\noindent\textbf{Datasets.} We evaluate OVSS on two flood segmentation datasets, Sen1Floods11~\cite{bonafilia2020sen1floods11} and WorldFloods~\cite{portales2023global}, and two land-cover datasets, DFC2020~\cite{9369830} and DynamicEarthNet~\cite{simumba2025geo,toker2022dynamicearthnet}, and ZSC on EuroSAT~\cite{lacoste2023geo,helber2019eurosat}, BigEarthNet~\cite{lacoste2023geo,sumbul2019bigearthnet}, and So2Sat~\cite{zhu2020so2sat, lacoste2023geo}, all using Sentinel-2 imagery. For RGB, we use two high-resolution aerial OVSS datasets, Chesapeake~\cite{lacoste2023geo,robinson2019large} and pv4ger~\cite{mayer20223d,lacoste2023geo}. Dataset details are in the Supplementary.

\noindent\textbf{Open-vocabulary semantic segmentation.} To evaluate the dense feature alignment of our method, we test it on MS and RGB OVSS datasets. For an MS image $\mathrm{\textbf{I}}\in\mathbb{R}^{C\times H \times W}$, we obtain backbone patch representations $p_\mathrm{s}^{\texttt{[p]}_2}(\phi(\mathrm{\textbf{I}})) = [\textbf{f}_1, ..., \textbf{f}_N]$. For textual representations, we encode prompts with the textual encoder $\psi(\texttt{text}) = \mathrm{\textbf{t}}$ and extract only the patch half $\mathrm{\textbf{t}} [D:2D]$. Finally, patch tokens are upsampled to the original image resolution $H \times W$; cosine similarity is computed for each pixel and each textual class representation $\mathrm{\textbf{t}}$, and the final class for each pixel is predicted with a softmax function. For RGB inputs, the process is the same, except we use the RGB patch projection head $p_\mathrm{s}^{\texttt{[p]}}$.

\noindent\textbf{Zero-shot classification.} To test global image understanding, we evaluate methods on zero-shot classification tasks. As in OVSS, we obtain $p_\mathrm{s}^{\texttt{[c]}_2/\texttt{[p]}_2}(\phi(\mathrm{\textbf{I}})) = [\textbf{c}, \textbf{f}_1, ..., \textbf{f}_N]$, where we also use the class token $\textbf{c}$.
The patch tokens are pooled and concatenated with $\textbf{c}$ to form a feature vector $[\textbf{c}; \mu (\textbf{f}_1, ..., \textbf{f}_N)]$ of dimension $2D$, which is compared to the entirety of $\psi(\texttt{text}) = \mathrm{\textbf{t}}$. For RGB inputs, we use $p_\mathrm{s}^{\texttt{[p]}/\texttt{[c]}}$ instead of $p_\mathrm{s}^{\texttt{[p]}_2/\texttt{[c]}_2}$.

\subsection{Results}
\label{subsec:main_results}

We report OVSS and ZSC results on 7 MS and 2 aerial RGB datasets in \Cref{tab:main}.
For our method, as well as \texttt{dino.txt}$\mathrm{_{LEO}}$ and \texttt{dino.txt}$\mathrm{_{LC}}$, we report the average over 3 seeds of image-text alignment, with standard deviations of at most 0.7 points for LEO. For Llama3-MS-CLIP~\cite{marimo2025beyond}, DOFA-CLIP~\cite{xiong2025dofa}, and LEO/LEO$\mathrm{_{LC}}$, we report both MS and RGB results.
The upper part of the table compares methods trained without auxiliary semantic guidance, while the lower part compares methods that include some form of it, such as land-cover segmentation or detection maps. Method details are in the Supplementary. 

Our method achieves state-of-the-art results in both settings on MS datasets. Without semantic guidance, LEO outperforms the next-best method, SkyCLIP~\cite{wang2024skyscript}, by 7.8 points, and even exceeds DOFA-CLIP~\cite{xiong2025dofa}, which uses semantic guidance, by 3.1 points; with land-cover guidance, LEO$_\text{LC}$ exceeds DOFA-CLIP by 7.7 points. LEO also outperforms \texttt{dino.txt}~\cite{jose2025dinov2} by 11.7 points, while LEO and LEO$_\text{LC}$ outperform its counterparts trained to match them, \texttt{dino.txt}$_\text{LEO}$ and \texttt{dino.txt}$_\text{LC}$, by 10.4 and 6.6 points, respectively. As \texttt{dino.txt} uses a single vision head for both class and patch tokens, \texttt{dino.txt}$_\text{LEO}$ fine-tunes this full head, which improves ZSC but degrades OVSS, consistent with \Cref{tab:ablation}. Notably, LEO achieves these results with a ViT-Base vision backbone, whereas all competing methods except Llama3-MS-CLIP~\cite{marimo2025beyond} use ViT-Large or larger. Llama3-MS-CLIP matches LEO in both backbone size and image-text training data, yet LEO outperforms it by 10.9 points. SATtxt~\cite{do2026spectrally} shares this training data and pairs a ViT-Large backbone with an 8B-parameter LLM text encoder, yet trails LEO by 11.3 points. Unlike related methods, we do not use high-resolution aerial imagery during image-text alignment to maintain a cleaner comparison, which reduces performance on high-resolution RGB aerial datasets. Land-cover guidance reduces this gap on Chesapeake.

\subsection{Multispectral vs.\ RGB}
\label{subsec:ms_vs_rgb}

To assess the benefit of additional spectral bands, we evaluate LEO and other MS methods on the same seven MS datasets using both MS and RGB-only inputs (\Cref{tab:main}), with average gains shown in \Cref{fig:first}. LEO benefits from MS inputs on every dataset, across both dense prediction and classification, with gains ranging from 2.0 to 8.4 points, while remaining competitive when given only RGB inputs. The gains are largest on flood segmentation: for floodwater that is not visibly distinct in RGB (\Cref{fig:qual}, first row), LEO recovers the full flooded area only with MS inputs. In contrast, Llama3-MS-CLIP~\cite{marimo2025beyond} performs similarly with either input (0.4 points higher with MS), while DOFA-CLIP~\cite{xiong2025dofa} performs 9.4 points worse with MS inputs, which may reflect its largely RGB-dominated image-text training data containing relatively few MS images~\cite{xiong2025dofa}. To rule out evaluation artifacts, we reproduce DOFA-CLIP's reported results on three datasets (see Supplementary).

\begin{table}[t]
\centering
\begin{tabular}{llcc}
\toprule
Visual backbone & Size & \textbf{MS} & \textbf{RGB} \\
\midrule
DINOv3~\cite{simeoni2025dinov3}            & ViT-L/16  & -   & 61.7 \\
RemoteCLIP~\cite{liu2024remoteclip}        & ViT-L/14  & -   & 59.0 \\
GeoRSCLIP~\cite{zhang2024rs5m}             & ViT-L/14  & -   & \bm1{62.2} \\
SkyCLIP~\cite{wang2024skyscript}           & ViT-L/14  & -   & 60.9 \\
SATtxt~\cite{do2026spectrally}             & ViT-L/16  & -   & 61.3 \\
Llama3-MS-CLIP~\cite{marimo2025beyond}     & ViT-B/16  & 60.0 & 59.1 \\
DOFA-CLIP~\cite{xiong2025dofa}             & SO400M/14 & \bm2{60.5} & 61.3 \\
\midrule
\rowcolor{oursblue}
\textbf{LEO}                               & ViT-B/16  & \bm1{67.4} & \bm2{62.0} \\
\bottomrule
\end{tabular}
\caption{\textbf{Linear probing results.} We evaluate linear probing on last-layer visual backbone features across seven MS datasets and report average results for MS and RGB-only inputs. \textcolor{goldD}{First} and \textcolor{silverD}{second} place results are marked.}
\label{tab:lp}
\end{table}

\subsection{Linear Probing}
\label{subsec:linear_probing}

To assess whether the lack of MS gains in existing methods comes from their visual backbones or their language alignment, we measure visual backbone expressiveness by training a linear probe on last-layer features across seven MS datasets and report the average results for MS and RGB-only inputs in \Cref{tab:lp}. Overall, our visual backbone outperforms others, achieving its best performance with MS inputs while matching the best other method on RGB inputs. Consistent with \Cref{fig:first}, our method is the only one that benefits from MS inputs: the visual backbones of Llama3-MS-CLIP~\cite{marimo2025beyond} and DOFA-CLIP~\cite{xiong2025dofa} show little improvement from MS inputs ($+$0.9 and $-$0.8 points), suggesting that they are ineffective at taking advantage of MS information. For DOFA-CLIP, language alignment further enlarges this gap to 9.4 points in favor of RGB inputs. By contrast, our backbone gains 5.4 points from MS inputs under linear probing, and LEO retains nearly all of this advantage after language alignment ($+$5.2 points on OVSS and ZSC), indicating that our alignment preserves the multispectral information in the visual backbone.

\newcommand{\up}[1]{\,{\tiny\textcolor{darkgreen}{$\uparrow$#1}}}
\newcommand{\dn}[1]{\,{\tiny\textcolor{red}{$\downarrow$#1}}}
\newcommand{\eq}{\,{\tiny\textcolor{darkgray}{$\pm$0.0}}}
\newcommand{\gup}[1]{\,{\tiny$\uparrow$#1}}
\newcommand{\gdn}[1]{\,{\tiny$\downarrow$#1}}
\begin{table}[t]
\centering
\resizebox{\linewidth}{!}{%
\setlength{\tabcolsep}{4pt}
\begin{tabular}{lcccc}
\toprule
 & \multicolumn{2}{c}{\textbf{MS}} & \multicolumn{2}{c}{\textbf{RGB}} \\
\cmidrule(lr){2-3} \cmidrule(lr){4-5}
 & OVSS & ZSC & OVSS & ZSC \\
\midrule
LEO(v) & 46.4 & 35.2 & 40.5 & 29.2 \\
\hspace{0.5em}\small\textcolor{darkgray}{$+$\texttt{[c]} head} & 46.4\eq & 47.1\up{11.9} & 40.5\eq & 29.2\eq \\
\hspace{1em}\small\textcolor{gray}{$+$\texttt{[p]} head}      & 39.1\dn{7.3} & 45.8\dn{1.3} & 40.5\eq & 29.2\eq \\
\rowcolor{oursblue}
\hspace{0.5em}\small\textcolor{darkgray}{$+$ RGB}        & 46.4\eq & 46.4\dn{0.7} & 40.5\eq & 42.1\up{12.9} \\
\rowcolor{oursblue}
\hspace{0.5em}\small\textcolor{gray}{$+$ LC maps} & 53.9\up{7.5} & 47.2\up{0.8} & 47.9\up{7.4} & 43.4\up{1.3} \\
\midrule
DEO~\cite{wolf2026deo} & 20.8 & 10.4 & 41.3 & 28.9 \\
\hspace{0.5em}\small\textcolor{darkgray}{$+$\texttt{[c]} head}  & 20.8\eq & 38.6\up{28.2} & 41.3\eq & 34.5\up{5.6} \\
\hspace{0.5em}\small\textcolor{darkgray}{$+$\texttt{[p]} head} & 33.9\up{13.1} & 41.0\up{2.4} & 32.9\dn{8.4} & 31.1\dn{3.4} \\
\hspace{0.5em}\small\textcolor{darkgray}{$+$ RGB}         & 36.6\up{2.7} & 41.3 \tiny\textcolor{darkgray}{$\uparrow$0.3} & 36.5\up{3.6} & 40.9\up{9.8} \\
\midrule
DINOv3~\cite{jose2025dinov2,simeoni2025dinov3} & - & - & 36.6 & 32.1 \\
\hspace{0.5em}\small\textcolor{darkgray}{$+$ vis. head} & - & - & 33.5\dn{3.1} & 39.3\up{7.2} \\
\hspace{0.5em}\small\textcolor{gray}{$+$ LC maps}      & - & - & 47.5\up{14.0} & 40.2\up{0.9} \\
\midrule
\midrule
\multicolumn{5}{l}{\textit{Trained components}} \\
\midrule
\rowcolor{oursblue}
LEO & \textbf{46.4} & 46.4 & 40.5 & 42.1 \\
\hspace{0.5em}\small\small\textcolor{orange}{{\faFire} $\psi$}\textcolor{cyan}{\hspace{1em}{\faSnowflake} $\phi$}        & 41.3\dn{5.1} & \textbf{47.4}\up{1.0} & 40.4 \tiny\textcolor{darkgray}{$\downarrow$0.1} & 40.8\dn{1.3} \\
\hspace{0.5em}\small\textcolor{cyan}{{\faSnowflake} $\psi$}\hspace{1em}\textcolor{orange}{{\faFire} $\phi$}     & 31.3\dn{15.1} & 46.2 \tiny\textcolor{darkgray}{$\downarrow$0.2} & 37.7\dn{2.8} & 42.6\up{0.5} \\
\hspace{0.5em}\small\textcolor{orange}{{\faFire} $\psi$\hspace{1em}{\faFire} $\phi$} & 45.7\dn{0.7} & 47.0\up{0.6} & \textbf{44.5}\up{4.0} & \textbf{44.1}\up{2.0} \\
\bottomrule
\end{tabular}
}
\caption{\textbf{Component ablation of image-text alignment.} We ablate components of our method using the LEO(v), DEO~\cite{wolf2026deo}, and DINOv3~\cite{simeoni2025dinov3} visual backbones with \texttt{dino.txt}~\cite{jose2025dinov2} across seven MS datasets and report average MS and RGB results on OVSS and ZSC. In the last block, we additionally train components beyond the vision-side projection heads. All numbers are 3-run averages. The full table is in the Supplementary.}
\label{tab:ablation}
\end{table}

\subsection{Ablation Study}
\label{subsec:ablation}

We conduct an ablation study of the components of our method, starting with the LEO(v) visual backbone paired with \texttt{dino.txt}, as shown in \Cref{tab:ablation}. We also perform the same ablation steps on DEO~\cite{wolf2026deo} and DINOv3~\cite{simeoni2025dinov3,jose2025dinov2}, the original visual backbone of \texttt{dino.txt}.

\noindent\textbf{Image-text alignment with projection heads.} We first perform image-text alignment by training the vision-side projection head $p_\mathrm{s}^{\texttt{[c]}_2}$ on MS EO image-caption pairs. For our method, this greatly improves MS ZSC performance by introducing a trainable translation layer that maps visual MS information to textual EO captions.
By allowing the patch projection head $p_\mathrm{s}^{\texttt{[p]}_2}$ to train as well, semantic segmentation performance is significantly reduced, as we rely solely on CLIP-style global alignment, which diminishes the local information captured by the pretrained patch heads. For DEO, training both projection heads aligns the visual-side MS information with the text encoder, thereby removing the collapse observed in its MS performance. However, it does not reach the performance of LEO(v) with trained heads. Training the vision head for DINOv3 has a similar effect to LEO(v), though to a smaller degree, since it is trained on RGB data, which it already interprets.

\noindent\textbf{Additional RGB training.} So far, image-text alignment has been performed only on MS imagery, and therefore only on the MS head $p_\mathrm{s}^{\texttt{[c]}_2}$. To preserve the RGB capabilities of our method acquired during visual pretraining, we train on RGB imagery 25\% of the time alongside the RGB projection head $p_\mathrm{s}^{\texttt{[c]}}$. This restores RGB capabilities, bringing them closer to our MS performance. A similar effect is observed for DEO. Training on more than 25\% RGB imagery does not improve performance (see Supplementary).

\noindent\textbf{Additional patch loss.} By adding the land-cover-derived patch loss described in \Cref{subsec:lca}, which trains patch heads $p_\mathrm{s}^{\texttt{[p]}}$ and $p_\mathrm{s}^{\texttt{[p]}_2}$, we significantly improve OVSS performance. A similar effect is observed for DINOv3, where its single vision head is trained.

\noindent\textbf{Pretraining vs.\ image-text alignment.} Applying the same image-text alignment to DEO, whose MS features are not aligned with DINOv3 during pretraining, also requires training its patch heads to remove the MS collapse. Even so, it trails our method on MS datasets, on both OVSS (36.6 vs.\ 46.4) and ZSC (41.3 vs.\ 46.4). Longer image-text alignment does not close this gap (see Supplementary), which we attribute to saturation on the comparatively small MS image-caption dataset. As unlabeled MS imagery is abundant while a sufficient number of MS image-caption pairs are scarce, aligning MS features with DINOv3 is best performed during visual pretraining (\Cref{subsec:ms_vision_backbone}).

\noindent\textbf{Trained component ablation.} During image-text alignment for LEO, we unlock only the vision-side projection heads. To motivate this choice, we additionally unlock the text encoder $\psi$, the visual encoder $\phi$, or both, and report results averaged over three seeds in the last block of Table~\ref{tab:ablation}. Unlocking only the visual encoder severely degrades dense performance, reducing MS OVSS by 15.1 points while leaving ZSC largely unchanged, because the global CLIP loss supervises only image-level features, allowing patch features of $\phi$ to drift. Unlocking only the text encoder reduces MS OVSS by 5.1 points and makes dense performance highly unstable across seeds (see Supplementary), as the text embeddings can drift away from the frozen patch features. Unlocking both encoders lets the two co-adapt, which nearly matches LEO on MS inputs ($-0.7$ OVSS, $+0.6$ ZSC) while improving RGB performance ($+4.0$ OVSS, $+2.0$ ZSC). This gain comes at the cost of training both encoders rather than only two projection heads. As it brings no benefit to MS inputs, which are the focus of this work, we keep both encoders frozen and note full unlocking as an option when RGB performance is the priority.

\begin{figure}[t]
  \centering
  \includegraphics[width=\linewidth]{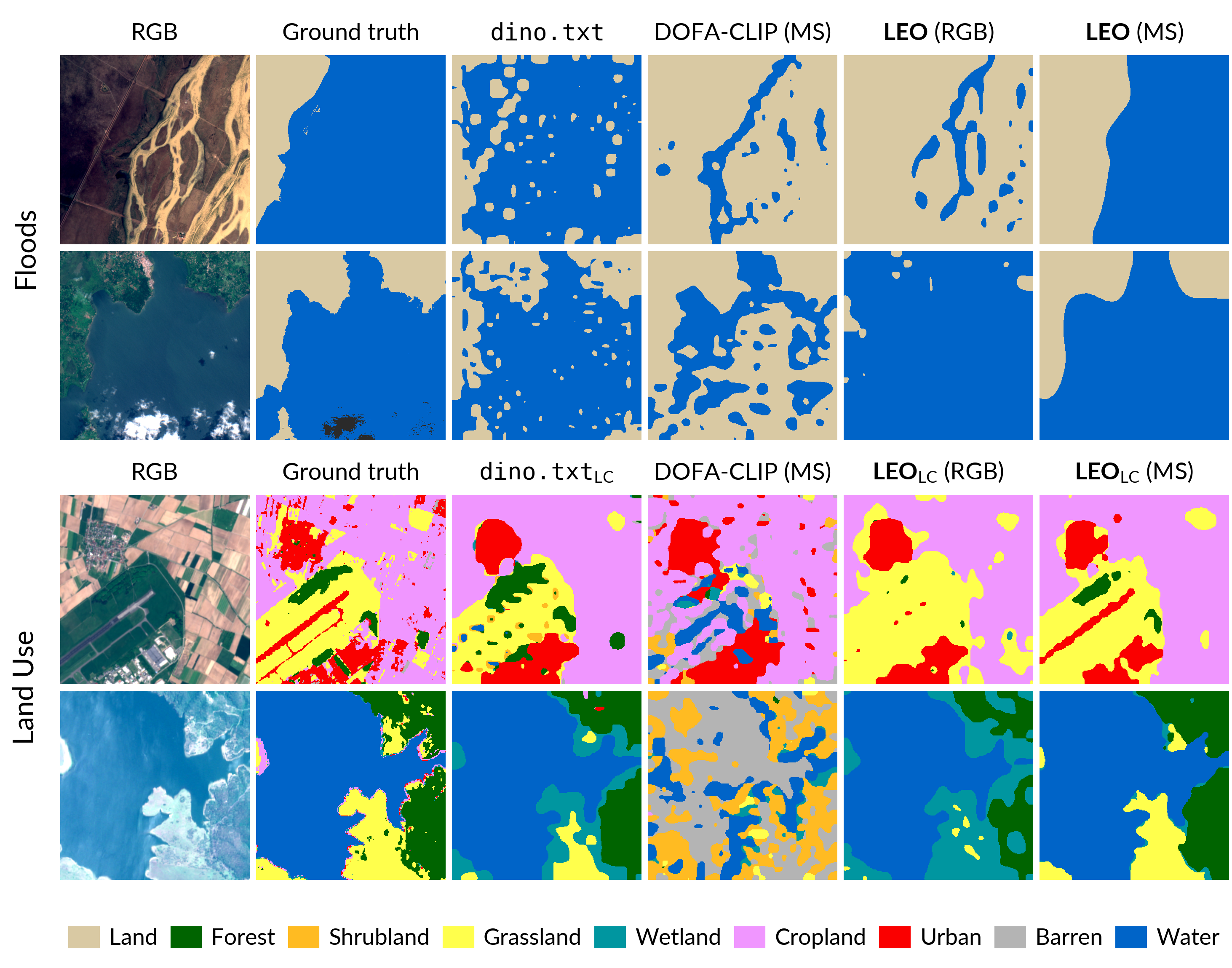}
  \caption{\textbf{Qualitative examples.} We show qualitative examples of OVSS for LEO and competing methods. For LEO, we compare results with both RGB and MS inputs. More qualitative results are in the Supplementary.}
  \label{fig:qual}
\end{figure}

\subsection{Qualitative results}

\Cref{fig:qual} shows qualitative OVSS results. For floods, \texttt{dino.txt} and DOFA-CLIP produce patchy predictions, whereas LEO’s are coherent. In the first example, where the floodwater is not visibly blue, LEO, with RGB inputs, segments only the visible water, whereas with MS inputs it recovers the full flooded area, which DOFA-CLIP misses despite using MS inputs. For land use, we show LEO$_\text{LC}$ and \texttt{dino.txt}$_\text{LC}$, as land-cover classes are what the land-cover guidance targets. LEO$_\text{LC}$ again yields more accurate, less patchy results, benefits from MS inputs over RGB, and outperforms DOFA-CLIP despite its semantic guidance.\\

\noindent\textbf{Limitations and future work.} LEO is limited by the availability of multispectral image-text data: existing datasets are far smaller than those used to train RGB VLMs, such as \texttt{dino.txt}, and their captions are derived from RGB information. LEO is also trained on low-resolution Sentinel-2 imagery, which limits its performance on high-resolution aerial data, and it does not yet support SAR inputs. In future work, we plan to address both issues and derive patch-level supervision directly from visual backbone features rather than from land-cover maps.
\section{Conclusion}

We presented \textbf{LEO}, a multispectral vision-language model for Earth observation. By aligning an MS visual backbone with DINOv3 during pretraining, we enable the frozen \texttt{dino.txt} text encoder to interpret MS features directly, leaving only vision-side projection heads to be trained on image-text pairs. LEO outperforms the next-best competing method trained without auxiliary semantic guidance by 7.8 points on MS datasets and, unlike existing MS EO VLMs, benefits from multispectral rather than RGB-only inputs by 5.2 points, while an optional land-cover patch loss, yielding LEO$_\mathrm{LC}$, further improves dense performance. These findings suggest that effective MS VLMs do not require large-scale image-text training or LLM-based text encoders but can be obtained by carefully aligning MS representations with existing language-aligned feature spaces.

\section*{LLM Usage Statement}
We use Claude (Anthropic) author-directed assistive tools for code generation and refinement, experimental planning, and results analysis. We use Claude and ChatGPT (OpenAI) for manuscript refinement, analysis, tables, and reviewer-style feedback. No LLMs or other generative AI tools were used by us for synthetic data generation, dataset creation, or annotation. All tool-assisted content was reviewed, substantially revised, and verified by the authors, who take full responsibility for the final manuscript and all scientific claims.

\clearpage
\setcounter{page}{1}
\maketitlesupplementary

\appendix

\renewcommand{\thetable}{S\arabic{table}}
\renewcommand{\thefigure}{S\arabic{figure}}

\section{Potential Negative Social Impact Statement}

LEO is an Earth Observation model and, as such, is primarily aimed at environmental tasks such as land-cover analysis, natural disaster damage assessment or prediction, temporal change analysis, and more. However, it can also be used for many negative purposes, such as military use or espionage. We do not endorse such uses. LEO is trained on Sentinel-2 imagery, which is of limited use in military applications due to its low spatial resolution of 10 m. In addition, its focus on multispectral data primarily targets tasks such as land-cover analysis, flood detection, and crop analysis. 

While LEO's image-text alignment is lightweight and uses minimal GPU resources, its pretraining stage is significantly more resource-intensive (\Cref{tab:gpu_hours}). Nevertheless, our model focuses on distillation from larger pretrained models and therefore requires less training on EO-specific data, reducing its final resource intensity.

\section{Architecture}

We provide more detail on the architecture used in the main paper for LEO.

\subsection{Projection Heads}
\label{subsec:p_heads}

All projection heads used for image-text alignment are 3-layer MLPs, similar to those in \cite{caron2021emerging}. We show the details in \Cref{tab:proj_heads}.

\begin{table}[h]
\centering
\small
\setlength{\tabcolsep}{4pt}
\resizebox{\linewidth}{!}{%
\begin{tabular}{llccc}
\toprule
Head & Branch & Layers & Dimensions & Stage 2 train \\
\midrule
\multicolumn{5}{l}{\textit{Class heads}} \\
\midrule
$p_\mathrm{s}^{\texttt{[c]}_2}$       & MS  & 3 & $768{\to}2048{\to}2048{\to}1024$ & LEO, LEO$\mathrm{_{LC}}$ \\
$p_\mathrm{s}^{\texttt{[c]}}$          & RGB & 3 & $768{\to}2048{\to}2048{\to}1024$ & LEO, LEO$\mathrm{_{LC}}$ \\
\midrule
\multicolumn{5}{l}{\textit{Patch heads}} \\
\midrule
$p_\mathrm{s}^{\texttt{[p]}_2}$  & MS  & 3 & $768{\to}2048{\to}2048{\to}1024$ & LEO$\mathrm{_{LC}}$ \\
$p_\mathrm{s}^{\texttt{[p]}}$      & RGB & 3 & $768{\to}2048{\to}2048{\to}1024$ & LEO$\mathrm{_{LC}}$ \\
\bottomrule
\end{tabular}
}
\caption{LEO projection head details.}
\label{tab:proj_heads}
\end{table}

\subsection{Visual Backbone}

Our visual backbone is the same as that used in DINOv3~\cite{simeoni2025dinov3}, which is a ViT-Base~\cite{dosovitskiy2020image} with RoPE~\cite{su2024roformer} positional embeddings and the added projection heads described in \Cref{subsec:p_heads}. During pretraining, we use additional projection heads, similar in structure to the described heads, for additional alignment, such as MS learning and regularization. We do not use the vision head from \texttt{dino.txt}~\cite{jose2025dinov2}.

\subsection{Textual Backbone}

We use the textual encoder from \texttt{dino.txt}~\cite{jose2025dinov2}, which is composed of 24 Transformer~\cite{vaswani2017attention} layers. In cases where it is trained, as in Table 4, we train only the last 12 layers to avoid catastrophic forgetting.

\subsection{Training Requirements}

We report training costs in GPU-hours in \Cref{tab:gpu_hours}.

\subsection{LEO Evaluation Details}

For the Sentinel-2 benchmarks, we use the imagery distributed with each dataset. The MS input uses ten bands in the order B4, B3, B2, B5, B6, B7, B8, B8A, B11, B12; the RGB input uses B4, B3, B2. Each band is standardized as $(x-\mu)/\sigma$ using the per-band statistics from the Sentinel-2 pretraining data; the RGB input uses the B4, B3, and B2 entries of the same statistics. The standardized image is then resized to the evaluation resolution using bicubic interpolation. MS and RGB inputs share the backbone and differ only in their patch embeddings and projection heads: the MS input goes through the MS patch embedding, and the RGB input goes through the RGB patch embedding.

\begin{table}[t]
\centering
\begin{tabular}{lc}
\toprule
Stage & GPU-hours \\
\midrule
Visual pretraining & 91.8 \\
Image-text alignment     & 0.5 \\
\midrule
Total                      & 92.3 \\
\bottomrule
\end{tabular}
\caption{Training costs in GPU-hours for visual pretraining and image-text alignment for LEO. We also provide the total. We use B300 GPUs for both training runs.}
\label{tab:gpu_hours}
\end{table}

\section{Datasets}

We show details for every dataset used in the main paper in \Cref{tab:dataset_details}.

\begin{table}[t]
\centering
\setlength{\tabcolsep}{4pt}
\begin{tabular}{llcccc}
\toprule
& & \multicolumn{2}{c}{\textbf{MS}} & \multicolumn{2}{c}{\textbf{RGB}} \\
\cmidrule(lr){3-4} \cmidrule(lr){5-6}
Model & Epochs & OVSS & ZSC & OVSS & ZSC \\
\midrule
\textbf{LEO} & 1 & 46.4 & \bm1{46.4} & 40.5 & \bm2{42.1} \\
 & 2 & 46.4 & \bm2{46.3} & 40.5 & 41.7 \\
 & 3 & 46.4 & 45.9 & 40.5 & \bm1{42.5} \\
 & 5 & 46.4 & 45.8 & 40.5 & 41.7 \\
\midrule
DEO~\cite{wolf2026deo} & 1 & \bm1{36.6} & \bm1{41.3} & \bm1{36.5} & \bm2{40.9} \\
 & 2 & \bm2{32.9} & \bm2{40.1} & \bm2{35.7} & 40.2 \\
 & 3 & 32.4 & 39.5 & 34.9 & 40.6 \\
 & 5 & 31.0 & 39.3 & 34.7 & \bm1{41.4} \\
\bottomrule
\end{tabular}
\caption{\textbf{Longer image-text alignment.} Results on the seven MS datasets when training for 1, 2, 3, or 5 epochs. LEO keeps its patch heads frozen, so only zero-shot classification can change; DEO trains both class and patch heads. One-epoch results are 3-run averages; longer runs are single runs.}
\label{tab:supp_longer_training}
\end{table}

\section{Additional Results}

We provide additional results to support the main paper, along with expanded tables from the main paper.

\subsection{Alignment Results}

We show the full alignment results (Table 1) in \Cref{tab:supp_dinov3_alignment}. We also show results for the original DEO~\cite{wolf2026deo}, which are very close to our retrained ViT-B variant.

\subsection{Main Results}

We report the mean and standard deviation for all trained methods from Table 2 over three runs with different seeds in \Cref{tab:supp_seed_variance}.

\subsection{Component Ablation}

We present the full component ablation table (Table 4), with averages and standard deviations over three runs, in \Cref{tab:supp_ablation_full}.

\subsection{Longer Image-Text Alignment}

In \Cref{tab:supp_longer_training}, we show results for longer image-text alignment for LEO and DEO. While longer training may slightly improve RGB ZSC, it largely degrades DEO's MS performance, especially OVSS, and slightly degrades LEO's MS ZSC performance.

\subsection{More RGB Training}

We ablate the amount of RGB training for LEO in \Cref{tab:supp_rgb_training}. We replace MS training steps with RGB steps in rows labeled \textit{Instead of MS}, and add additional RGB steps in rows labeled \textit{In addition to MS}. OVSS results do not change because we do not train the patch head. Training on more than 25\% of the steps in both cases slightly increases RGB ZSC with no consistent gain in MS ZSC. Since our focus is on MS performance and keeping compute equal, we train on 25\% of the steps that replace MS steps.

\begin{table}[t]
\centering
\setlength{\tabcolsep}{4pt}
\begin{tabular}{llcccc}
\toprule
& & \multicolumn{2}{c}{MS} & \multicolumn{2}{c}{RGB} \\
\cmidrule(lr){3-4} \cmidrule(lr){5-6}
\textbf{LEO} & RGB steps  & OVSS & ZSC & OVSS & ZSC \\
\midrule
\multirow{4}{1.4cm}{\raggedright\textit{Instead of MS}}
  & none & 46.4 & \bm1{47.1} & 40.5 & 29.2 \\
  & 25\% & 46.4 & 46.4 & 40.5 & 42.1 \\
  & 50\% & 46.4 & 46.5 & 40.5 & 42.8 \\
  & 75\% & 46.4 & 46.2 & 40.5 & \bm1{43.4} \\
\midrule
\multirow{4}{1.4cm}{\raggedright\textit{In addition to MS}}
  & 25\%  & 46.4 & \bm2{46.8} & 40.5 & 42.0 \\
  & 50\%  & 46.4 & 45.8 & 40.5 & 43.0 \\
  & 75\%  & 46.4 & 45.7 & 40.5 & 43.0 \\
  & 100\% & 46.4 & 45.5 & 40.5 & \bm2{43.1} \\
\bottomrule
\end{tabular}
\caption{\textbf{RGB training ablation.} Averages across the seven MS OVSS and ZSC datasets. LEO keeps its patch heads frozen during image-text alignment, so its OVSS cannot change.}
\label{tab:supp_rgb_training}
\end{table}

\begin{table}[t]
\centering
\small
\begin{tabular}{lcccc}
\toprule
& \multicolumn{2}{c}{In~\cite{xiong2025dofa}} & \multicolumn{2}{c}{Ours} \\
\cmidrule(lr){2-3} \cmidrule(lr){4-5}
Dataset & RGB & MS & RGB & MS \\
\midrule
EuroSAT~\cite{lacoste2023geo,helber2019eurosat} & \multicolumn{2}{c}{59.0*} & 78.8 & 58.5 \\
So2Sat~\cite{zhu2020so2sat, lacoste2023geo}  & 18.0 & 14.6 & 25.4 & 14.9 \\
BigEarthNet~\cite{sumbul2019bigearthnet, lacoste2023geo} & - & 23.7 & 28.5 & 24.9 \\
\bottomrule
\end{tabular}
\caption{\textbf{Comparison with published DOFA-CLIP results.} Zero-shot classification reported in the DOFA-CLIP paper vs. our evaluation. EuroSAT and So2Sat report Top-1 accuracy. For BigEarthNet, we report micro-F1 using the decision threshold selected on the validation set. *input type is not specified for EuroSAT in~\cite{xiong2025dofa}.}
\label{tab:supp_dofa_repro}
\end{table}

\subsection{Comparison with DOFA-CLIP Results}

We compare our DOFA-CLIP results with the published scores from~\cite{xiong2025dofa} (\Cref{tab:supp_dofa_repro}). In our evaluation, DOFA-CLIP-L-384 (SigLIP SO400M/14 backbone) scores higher with RGB than with MS inputs on EuroSAT, So2Sat, and BigEarthNet. Our MS scores differ from the published values by 0.5, 0.3, and 1.2 points, respectively. The published So2Sat and BigEarthNet scores are based on the smaller DOFA-CLIP-B-224. On So2Sat, the DOFA-CLIP authors likewise report higher accuracy with RGB than with five Sentinel-2 bands. Thus, the RGB advantage on So2Sat appears in both evaluations, despite the difference in model size. For EuroSAT, the authors use the same large model as we do but do not specify the input modality. They evaluate on the 2,700-image test split of the original dataset~\cite{helber2019eurosat}, whereas we use the 1,000-image test split from GEO-Bench~\cite{lacoste2023geo}. \Cref{tab:supp_method_details} describes our input normalization and dense feature extraction.

\begin{table*}[t]
\centering
\setlength{\tabcolsep}{4pt}
\resizebox{\linewidth}{!}{%
\begin{tabular}{llcccccccl}
\toprule
Dataset & In paper & Image Size & \# Classes & Train & Validation & Test & \# Bands & GSD (m) & Sensors \\
\midrule
\multicolumn{5}{l}{\textit{Open-vocabulary semantic segmentation}} \\
\midrule
Sen1Floods11~\cite{bonafilia2020sen1floods11}                       & S1F11   & 512$\times$512 & 2  & 252  & 89   & 90   & 13 & 10.0 & Sentinel-2 \\
WorldFloods v2~\cite{portales2023global}                            & WF      & 512$\times$512 & 2  & 606$^\dagger$ & 368$^\dagger$ & 403$^\dagger$ & 13 & 10.0 & Sentinel-2 \\
DFC2020~\cite{9369830}                                              & DFC     & 256$\times$256 & 8  & -$^\ddagger$ & 986 & 5128 & 13 & 10.0 & Sentinel-2 \\
DynamicEarthNet~\cite{toker2022dynamicearthnet,simumba2025geo}      & DEN     & 512$\times$512 & 7  & 700  & 100  & 200  & 12 & 10.0 & Sentinel-2 \\
m-chesapeake-landcover~\cite{robinson2019large,lacoste2023geo}      & Chesa.  & 256$\times$256 & 6  & 3000 & 1000 & 1000 & 3  & 1.0  & NAIP (RGB) \\
m-pv4ger-seg~\cite{mayer20223d,lacoste2023geo}                      & pv4ger  & 320$\times$320 & 2  & 3000 & 403  & 403  & 3  & 0.1  & Aerial (RGB) \\
\midrule
\midrule
\multicolumn{5}{l}{\textit{Zero-shot classification}} \\
\midrule
m-eurosat~\cite{helber2019eurosat,lacoste2023geo}                   & EuroSAT & 64$\times$64   & 10 & 2000  & 1000 & 1000 & 13 & 10.0 & Sentinel-2 \\
m-bigearthnet~\cite{sumbul2019bigearthnet,lacoste2023geo}           & BEN     & 120$\times$120 & 43 & 20000 & 1000 & 1000 & 12 & 10.0 & Sentinel-2 \\
m-so2sat~\cite{zhu2020so2sat,lacoste2023geo}                        & So2Sat  & 32$\times$32   & 17 & 19992 & 986  & 986  & 10 & 10.0 & Sentinel-2 \\
\bottomrule
\end{tabular}
}
\caption{\textbf{Dataset details.} \# (number) of classes counts the scored classes; all other pixels are ignored. DFC2020 has 10 classes; savanna and snow are ignored, so it lists 8. Chesapeake has 7 labels, with \textit{no-data} ignored, so it lists 6. For Sen1Floods11, we ignore clouds. Train splits are used only for linear probing; zero-shot and open-vocabulary results use the validation split to select the input resolution and report final results on the test split. $^\dagger$WorldFloods comes as large flood event scenes. We cut 512$\times$512-pixel tiles from the scenes and keep a tile only if it is at least 50\% cloud-free and labeled. For validation and testing, we randomly sample at most 60 tiles per scene, resulting in 368 and 403 tiles. For training (used only for linear probing), we instead randomly sample 60 scenes (due to the large number of scenes), and from each scene, at most 20 tiles, resulting in 606 tiles. $^\ddagger$No public train split; linear probing uses 80\% of the validation tiles for training and the remaining 20\% for validation.}
\label{tab:dataset_details}
\end{table*}

\begin{table*}[t]
\centering
\begin{tabular}{llccccccccc}
\toprule
& & \multicolumn{5}{c}{OVSS} & \multicolumn{4}{c}{Zero-shot classification} \\
\cmidrule(lr){3-7} \cmidrule(lr){8-11}
Vision backbone & Input & S1F11 & WF & DFC & DEN & \textit{Avg} & EuroSAT & BEN & So2Sat & \textit{Avg} \\
\midrule
DINOv3~\cite{simeoni2025dinov3,jose2025dinov2} & RGB & 52.6 & 56.3 & 24.1 & 13.4 & 36.6 & 56.9 & 25.2 & 14.3 & 32.1 \\
\midrule
\multirow{2}{*}{DEO~\cite{wolf2026deo} (Swin-B)} & RGB & 55.1 & 64.1 & 24.6 & 7.4 & 37.8 & 52.4 & 19.0 & 15.6 & 29.0 \\
                                                 & MS  & 43.8 & 33.6 & 5.2 & 3.1 & 21.4 & 16.2 & 10.5 & 6.2 & 11.0 \\
\multirow{2}{*}{DEO (ViT-B)}                     & RGB & 63.1 & 64.7 & 28.0 & 9.4 & 41.3 & 53.5 & 19.1 & 14.1 & 28.9 \\
                                                 & MS  & 43.8 & 33.6 & 3.9 & 1.9 & 20.8 & 15.3 & 11.2 & 4.6 & 10.4 \\
\midrule
\multirow{2}{*}{LEO(v)}                          & RGB & 60.3 & 65.4 & 26.7 & 9.6 & 40.5 & 54.2 & 19.0 & 14.3 & 29.2 \\
                                                 & MS  & 68.7 & 71.8 & 28.8 & 16.1 & 46.4 & 65.4 & 21.6 & 18.7 & 35.2 \\
\bottomrule
\end{tabular}
\caption{\textbf{Full Table 1 results.}}
\label{tab:supp_dinov3_alignment}
\end{table*}

\subsection{Linear Probing}

For the linear probing experiments in Table 3, all backbones are kept frozen.
For segmentation, the probe is a BatchNorm layer followed by a $1\times1$ convolution on the last-layer patch tokens (after the final normalization). Its logits are bilinearly upsampled to the label resolution. It is trained with cross-entropy for 10 epochs with a batch size of 16. The input size is selected on validation from \{224, 448\}.

For classification, the features are standardized using the training-set mean and standard deviation and then fed into a single linear layer. It is trained for 3000 steps with a batch size of 256, using cross-entropy or binary cross-entropy for the multi-label BigEarthNet; the input size is 224. For our models, the classification feature is the CLS token. For competitors, we use their backbone tokens. All probes use AdamW with a weight decay of $10^{-4}$ and a one-cycle learning-rate schedule with 10\% warm-up, without data augmentation. The learning rate is selected from $\{10^{-3}, 3\cdot10^{-3}, 10^{-2}\}$ together with the input size on the validation split, and the selected probe is evaluated once on the test split. DOFA-CLIP is probed at its native 384\,px.

\section{Method Details}

We provide details on the methods we compare against in \Cref{tab:supp_method_details}. We report their sources, visual and textual encoders, input-handling methods, and how we extract dense features.

\section{Additional Qualitative Results}

\begin{figure*}[t]
  \centering
  \includegraphics[width=\textwidth]{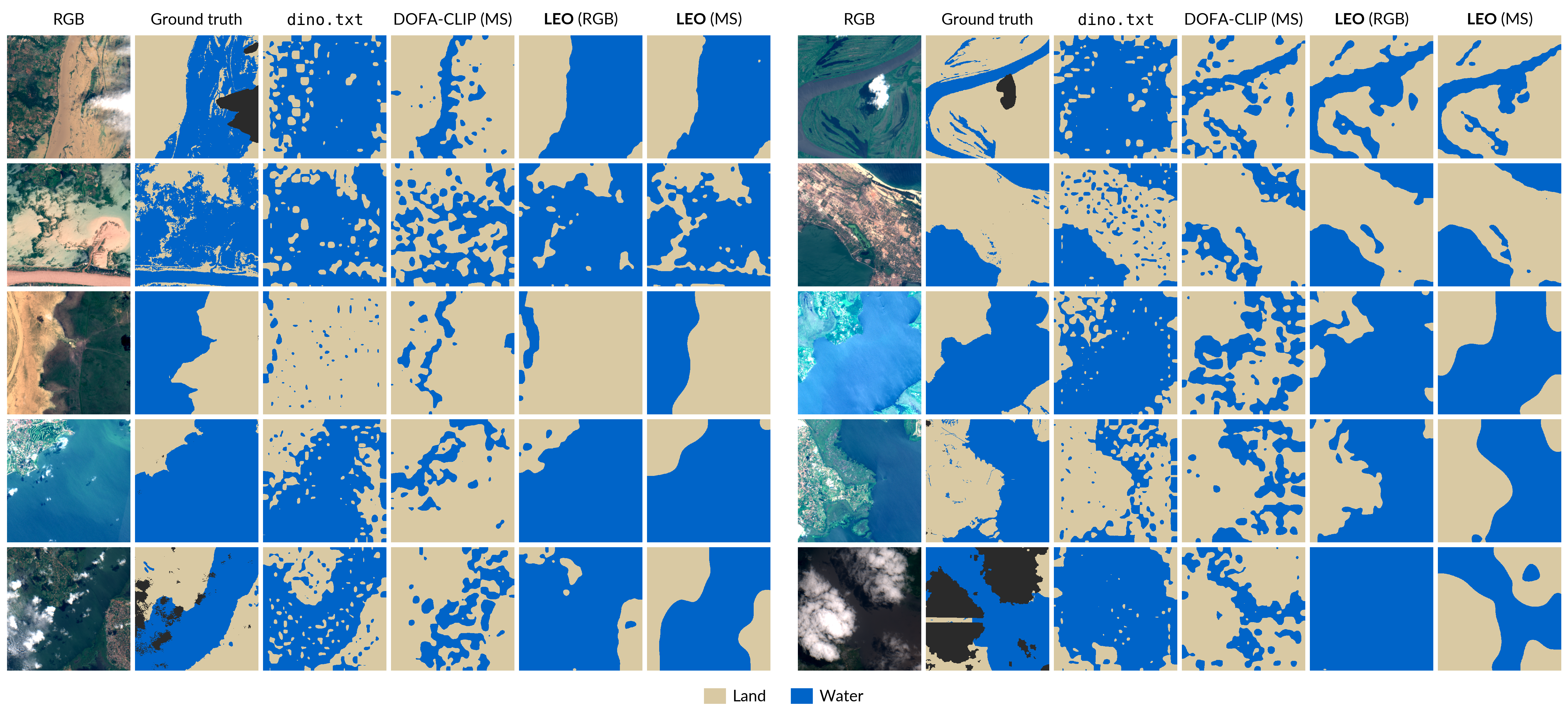}
  \caption{\textbf{Additional qualitative examples for floods.} Examples were selected from Sen1Floods11~\cite{bonafilia2020sen1floods11} and WorldFloods v2~\cite{portales2023global}.}
  \label{fig:qual_floods}
\end{figure*}

\begin{figure*}[t]
  \centering
  \includegraphics[width=\textwidth]{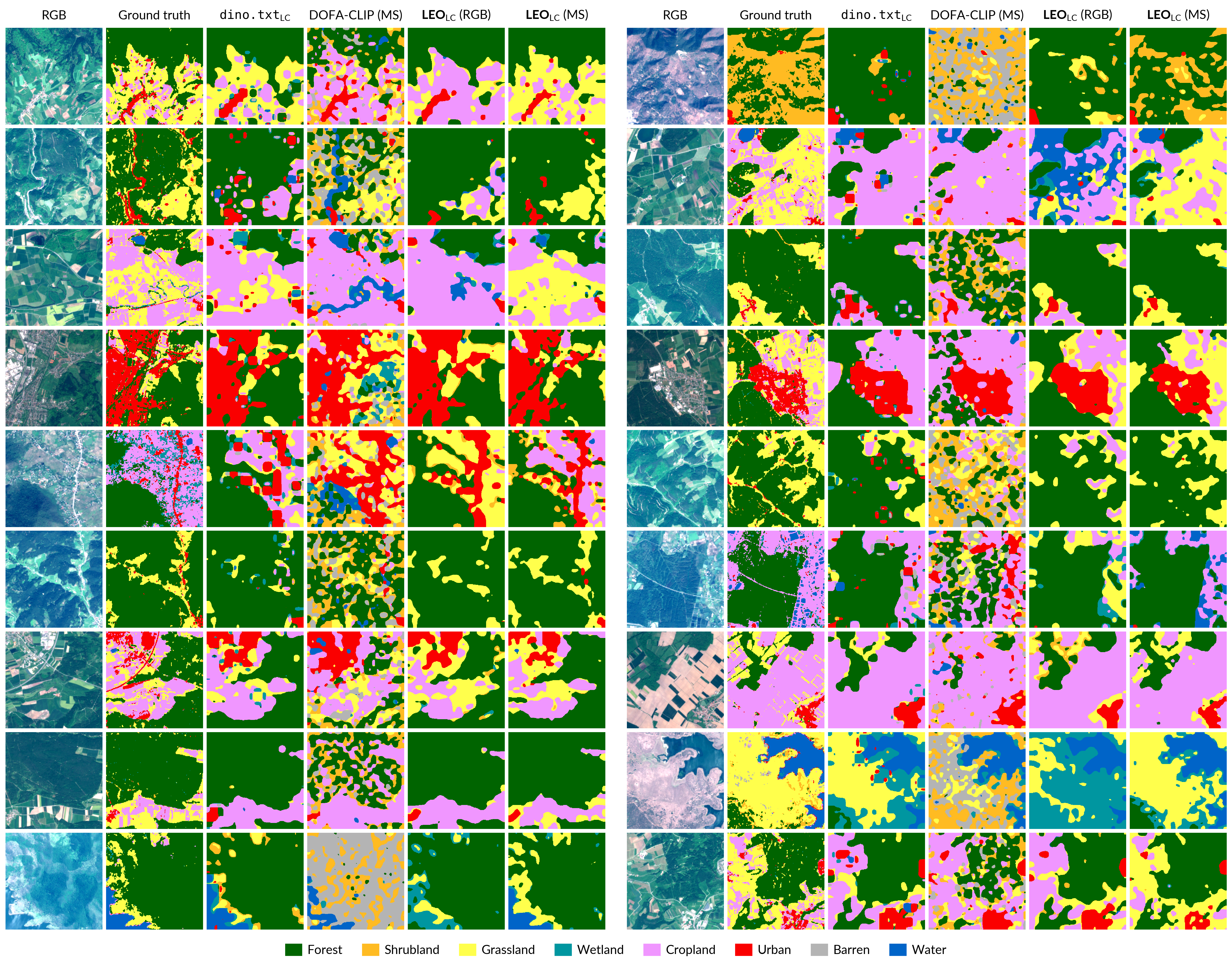}
  \caption{\textbf{Additional qualitative examples for land cover.} Examples selected from DFC2020~\cite{9369830}.}
  \label{fig:qual_crop}
\end{figure*}

In \Cref{fig:qual_floods,fig:qual_crop}, we show additional qualitative examples of floods and land cover.

\begin{table*}[t]
\centering
\newcommand{\pmstd}[1]{\,{\scriptsize$\pm$#1}}
\setlength{\tabcolsep}{4pt}
\resizebox{\linewidth}{!}{%
\begin{tabular}{llcc|cccccccc}
\toprule
& & \multicolumn{2}{c}{\textbf{RGB}} & \multicolumn{8}{c}{\textbf{MS}} \\
\cmidrule(lr){3-4} \cmidrule(lr){5-12}
& & \multicolumn{2}{c}{OVSS} & \multicolumn{4}{c}{OVSS} & \multicolumn{3}{c}{Zero-shot classification} & \\
\cmidrule(lr){3-4} \cmidrule(lr){5-8} \cmidrule(lr){9-11}
Method & Input & Chesa. & pv4ger & S1F11 & WF & DFC & DEN & EuroSAT & BEN & So2Sat & \textit{Avg} \\
\midrule
\texttt{dino.txt}$\mathrm{_{LEO}}$ & RGB & 32.5\pmstd{0.4} & 62.8\pmstd{0.9} & 44.4\pmstd{1.1} & 47.7\pmstd{1.5} & 28.1\pmstd{0.4} & 13.8\pmstd{0.1} & 70.1\pmstd{0.3} & 23.7\pmstd{0.1} & 24.1\pmstd{1.0} & 36.0\pmstd{0.1} \\
\midrule
\multirow{2}{*}{\textbf{LEO}} & RGB & 26.6\pmstd{0.0} & 67.3\pmstd{0.0} & 60.3\pmstd{0.0} & 65.4\pmstd{0.0} & 26.7\pmstd{0.0} & 9.6\pmstd{0.0} & 75.2\pmstd{0.4} & 24.8\pmstd{0.2} & 26.5\pmstd{0.4} & 41.2\pmstd{0.1} \\
                              & MS  & - & - & 68.7\pmstd{0.0} & 71.8\pmstd{0.0} & 28.8\pmstd{0.0} & 16.1\pmstd{0.0} & 81.3\pmstd{0.3} & 26.8\pmstd{0.2} & 31.1\pmstd{0.7} & 46.4\pmstd{0.1} \\
\midrule
\midrule
\multicolumn{12}{l}{\textit{Auxiliary semantic guidance}} \\
\midrule
\texttt{dino.txt}$\mathrm{_{LC}}$ & RGB & 42.6\pmstd{0.2} & 67.6\pmstd{0.7} & 60.9\pmstd{0.9} & 65.0\pmstd{0.1} & 40.7\pmstd{0.2} & 23.4\pmstd{0.2} & 71.0\pmstd{0.2} & 24.2\pmstd{0.1} & 25.5\pmstd{0.8} & 44.4\pmstd{0.1} \\
\midrule
\multirow{2}{*}{\textbf{LEO}$\mathrm{_{LC}}$} & RGB & 40.1\pmstd{0.5} & 61.3\pmstd{0.5} & 61.1\pmstd{0.5} & 65.6\pmstd{0.6} & 39.8\pmstd{0.2} & 25.4\pmstd{0.1} & 78.2\pmstd{0.1} & 25.3\pmstd{0.2} & 26.7\pmstd{0.6} & 46.0\pmstd{0.1} \\
                                              & MS  & - & - & 71.5\pmstd{0.4} & 74.0\pmstd{0.3} & 43.3\pmstd{0.2} & 26.6\pmstd{0.7} & 81.2\pmstd{0.3} & 27.3\pmstd{0.0} & 33.3\pmstd{0.3} & 51.0\pmstd{0.1} \\
\bottomrule
\end{tabular}
}
\caption{\textbf{Mean and standard deviation over three training seeds.} We show results for the trained models in Table 2. LEO's segmentation results have zero variance because its patch projection heads remain frozen during image-text alignment; only the class-token heads, and therefore only zero-shot classification, vary between runs.}
\label{tab:supp_seed_variance}
\end{table*}

\newcommand{\twol}[1]{\begin{tabular}[t]{@{}l@{}}#1\end{tabular}}
\newcommand{\gh}[2]{\twol{\href{https://github.com/#1/#2}{\texttt{github.com/}}\\\href{https://github.com/#1/#2}{\texttt{#1/}}\\\href{https://github.com/#1/#2}{\texttt{#2}}}}
\begin{table*}[t]
\centering
\resizebox{\linewidth}{!}{%
\setlength{\tabcolsep}{4pt}
\begin{tabular}{lllll}
\toprule
Model & Backbone (image / text) & Code & Normalization & OVSS \\
\midrule
dino.txt~\cite{jose2025dinov2} &
\twol{DINOv3 ViT-L/16\\+ \texttt{dino.txt} vis. head \\+ \texttt{dino.txt} text enc.} &
\gh{facebookresearch}{dinov3} &
\twol{$\lfloor\mathrm{clip}(x/2000\cdot255,0,255)\rfloor$,\\ImageNet mean/std} &
Supported \\
SATtxt~\cite{do2026spectrally} &
\twol{DINOv3 ViT-L/16\\+ Llama-3-8B} &
\gh{ikhado}{sattxt} &
\twol{$\mathrm{clip}(x/2000,0,1)$,\\CLIP mean/std} &
Supported \\
RemoteCLIP~\cite{liu2024remoteclip} &
\twol{CLIP ViT-L/14\\+ CLIP text enc.} &
\gh{ChenDelong1999}{RemoteCLIP} &
\twol{$\lfloor\mathrm{clip}(x/2000\cdot255,0,255)\rfloor$,\\CLIP mean/std} &
MaskCLIP~\cite{zhou2022extract} \\
GeoRSCLIP~\cite{zhang2024rs5m} &
\twol{CLIP ViT-L/14\\+ CLIP text enc.} &
\gh{om-ai-lab}{RS5M} &
As RemoteCLIP &
MaskCLIP~\cite{zhou2022extract} \\
SkyCLIP~\cite{wang2024skyscript} &
\twol{CLIP ViT-L/14\\+ CLIP text enc.} &
\gh{wangzhecheng}{SkyScript} &
As RemoteCLIP &
MaskCLIP~\cite{zhou2022extract} \\
Llama3-MS-CLIP~\cite{marimo2025beyond} &
\twol{CLIP ViT-B/16\\+ CLIP text enc.} &
\gh{IBM}{MS-CLIP} &
\twol{MS: per-band mean/std\\as in the repository;\\RGB: B4, B3, B2, rest at mean} &
MaskCLIP~\cite{zhou2022extract} \\
DOFA-CLIP~\cite{xiong2025dofa} &
\twol{SigLIP SO400M/14\\+ SigLIP text enc.} &
\gh{xiong-zhitong}{DOFA-CLIP} &
\twol{$\mathrm{clip}(x/3000,0,1)$,\\then $(x-0.5)/0.5$;\\wavelengths as in the repository} &
\twol{MAP pooling head\\on every patch token} \\
\bottomrule
\end{tabular}
}
\caption{\textbf{Evaluated methods.} All models are used frozen, with their released weights. $x$ denotes inputs.}
\label{tab:supp_method_details}
\end{table*}

\begin{table*}[t]
\centering
\newcommand{\pmstd}[1]{\,{\scriptsize$\pm$#1}}
\setlength{\tabcolsep}{4pt}
\begin{tabular}{llccccccc}
\toprule
& & \multicolumn{4}{c}{OVSS} & \multicolumn{3}{c}{Zero-shot classification} \\
\cmidrule(lr){3-6} \cmidrule(lr){7-9}
 & Input & S1F11 & WF & DFC & DEN & EuroSAT & BEN & So2Sat \\
\midrule
LEO(v) & RGB & 60.3 & 65.4 & 26.7 & 9.6 & 54.2 & 19.0 & 14.3 \\
 & MS & 68.7 & 71.8 & 28.8 & 16.1 & 65.4 & 21.6 & 18.7 \\
\hspace{0.5em}\small\textcolor{darkgray}{$+$\texttt{[c]} head} & RGB & 60.3\pmstd{0.0} & 65.4\pmstd{0.0} & 26.7\pmstd{0.0} & 9.6\pmstd{0.0} & 54.3\pmstd{0.1} & 19.0\pmstd{0.0} & 14.3\pmstd{0.0} \\
 & MS & 68.7\pmstd{0.0} & 71.8\pmstd{0.0} & 28.8\pmstd{0.0} & 16.1\pmstd{0.0} & 82.5\pmstd{0.8} & 26.9\pmstd{0.1} & 31.8\pmstd{0.4} \\
\hspace{1em}\small\textcolor{gray}{$+$\texttt{[p]} head} & RGB & 60.3\pmstd{0.0} & 65.4\pmstd{0.0} & 26.7\pmstd{0.0} & 9.6\pmstd{0.0} & 54.3\pmstd{0.1} & 19.0\pmstd{0.0} & 14.3\pmstd{0.0} \\
 & MS & 53.6\pmstd{0.4} & 60.9\pmstd{1.1} & 24.0\pmstd{0.5} & 17.8\pmstd{0.4} & 79.1\pmstd{0.2} & 26.8\pmstd{0.0} & 31.6\pmstd{0.3} \\
\hspace{0.5em}\small\textcolor{darkgray}{$+$ RGB} & RGB & 60.3\pmstd{0.0} & 65.4\pmstd{0.0} & 26.7\pmstd{0.0} & 9.6\pmstd{0.0} & 75.2\pmstd{0.4} & 24.8\pmstd{0.2} & 26.5\pmstd{0.4} \\
 & MS & 68.7\pmstd{0.0} & 71.8\pmstd{0.0} & 28.8\pmstd{0.0} & 16.1\pmstd{0.0} & 81.3\pmstd{0.3} & 26.8\pmstd{0.2} & 31.1\pmstd{0.7} \\
\hspace{0.5em}\small\textcolor{gray}{$+$ LC maps} & RGB & 61.1\pmstd{0.5} & 65.6\pmstd{0.6} & 39.8\pmstd{0.2} & 25.4\pmstd{0.1} & 78.2\pmstd{0.1} & 25.3\pmstd{0.2} & 26.7\pmstd{0.6} \\
 & MS & 71.5\pmstd{0.4} & 74.0\pmstd{0.3} & 43.3\pmstd{0.2} & 26.6\pmstd{0.7} & 81.2\pmstd{0.3} & 27.3\pmstd{0.0} & 33.3\pmstd{0.3} \\
\midrule
DEO~\cite{wolf2026deo} & RGB & 63.1 & 64.7 & 28.0 & 9.4 & 53.5 & 19.1 & 14.1 \\
 & MS & 43.8 & 33.6 & 3.9 & 1.9 & 15.3 & 11.2 & 4.6 \\
\hspace{0.5em}\small\textcolor{darkgray}{$+$\texttt{[c]} head} & RGB & 63.1\pmstd{0.0} & 64.7\pmstd{0.0} & 28.0\pmstd{0.0} & 9.4\pmstd{0.0} & 62.3\pmstd{0.5} & 20.8\pmstd{0.2} & 20.2\pmstd{0.3} \\
 & MS & 43.8\pmstd{0.0} & 33.6\pmstd{0.0} & 3.9\pmstd{0.0} & 1.9\pmstd{0.0} & 69.6\pmstd{0.4} & 21.1\pmstd{0.1} & 25.3\pmstd{0.3} \\
\hspace{0.5em}\small\textcolor{darkgray}{$+$\texttt{[p]} head} & RGB & 43.9\pmstd{0.7} & 56.1\pmstd{0.4} & 22.5\pmstd{1.5} & 9.0\pmstd{0.3} & 52.0\pmstd{0.4} & 20.7\pmstd{0.2} & 20.6\pmstd{1.0} \\
 & MS & 44.2\pmstd{0.8} & 47.7\pmstd{1.8} & 29.0\pmstd{0.2} & 14.8\pmstd{0.5} & 73.2\pmstd{0.5} & 22.8\pmstd{0.1} & 27.1\pmstd{0.9} \\
\hspace{0.5em}\small\textcolor{darkgray}{$+$ RGB} & RGB & 52.2\pmstd{0.6} & 54.8\pmstd{1.4} & 25.7\pmstd{1.7} & 13.4\pmstd{1.4} & 75.7\pmstd{1.1} & 23.6\pmstd{0.3} & 23.5\pmstd{0.8} \\
 & MS & 49.0\pmstd{1.3} & 51.6\pmstd{3.1} & 30.0\pmstd{0.4} & 16.0\pmstd{0.9} & 73.3\pmstd{0.4} & 22.6\pmstd{0.3} & 28.0\pmstd{0.9} \\
\midrule
DINOv3~\cite{jose2025dinov2,simeoni2025dinov3} & RGB & 52.6 & 56.3 & 24.1 & 13.4 & 56.9 & 25.2 & 14.3 \\
\hspace{0.5em}\small\textcolor{darkgray}{$+$ vis. head} & RGB & 44.4\pmstd{1.1} & 47.7\pmstd{1.5} & 28.1\pmstd{0.4} & 13.8\pmstd{0.1} & 70.1\pmstd{0.3} & 23.7\pmstd{0.1} & 24.1\pmstd{1.0} \\
\hspace{0.5em}\small\textcolor{gray}{$+$ LC maps} & RGB & 60.9\pmstd{0.9} & 65.0\pmstd{0.1} & 40.7\pmstd{0.2} & 23.4\pmstd{0.2} & 71.0\pmstd{0.2} & 24.2\pmstd{0.1} & 25.5\pmstd{0.8} \\
\midrule
\midrule
\multicolumn{9}{l}{\textit{Trained components}} \\
\midrule
LEO & RGB & 60.3\pmstd{0.0} & 65.4\pmstd{0.0} & 26.7\pmstd{0.0} & 9.6\pmstd{0.0} & 75.2\pmstd{0.4} & 24.8\pmstd{0.2} & 26.5\pmstd{0.4} \\
 & MS & 68.7\pmstd{0.0} & 71.8\pmstd{0.0} & 28.8\pmstd{0.0} & 16.1\pmstd{0.0} & 81.3\pmstd{0.3} & 26.8\pmstd{0.2} & 31.1\pmstd{0.7} \\
\hspace{0.5em}\small\textcolor{orange}{{\faFire} $\psi$}\hspace{1em}\textcolor{cyan}{{\faSnowflake} $\phi$} & RGB & 61.2\pmstd{8.6} & 58.9\pmstd{6.1} & 27.9\pmstd{0.7} & 13.5\pmstd{1.0} & 72.6\pmstd{0.8} & 22.7\pmstd{0.6} & 27.1\pmstd{1.6} \\
 & MS & 60.5\pmstd{6.5} & 57.2\pmstd{7.2} & 30.4\pmstd{1.1} & 17.3\pmstd{3.1} & 80.3\pmstd{0.3} & 23.6\pmstd{0.3} & 38.2\pmstd{0.9} \\
\hspace{0.5em}\small\textcolor{cyan}{{\faSnowflake} $\psi$}\hspace{1em}\textcolor{orange}{{\faFire} $\phi$} & RGB & 52.1\pmstd{2.1} & 56.5\pmstd{1.1} & 27.5\pmstd{0.8} & 14.7\pmstd{1.5} & 76.7\pmstd{1.8} & 25.1\pmstd{0.5} & 26.0\pmstd{1.3} \\
 & MS & 37.2\pmstd{3.5} & 47.5\pmstd{2.0} & 23.9\pmstd{0.8} & 16.6\pmstd{1.0} & 81.2\pmstd{0.9} & 29.3\pmstd{0.3} & 28.1\pmstd{0.9} \\
\hspace{0.5em}\small\textcolor{orange}{{\faFire} $\psi$\hspace{1em}{\faFire} $\phi$} & RGB & 65.3\pmstd{1.0} & 60.5\pmstd{0.3} & 33.2\pmstd{0.2} & 19.3\pmstd{0.7} & 74.1\pmstd{1.5} & 24.7\pmstd{0.9} & 33.4\pmstd{0.5} \\
 & MS & 65.3\pmstd{2.5} & 61.6\pmstd{1.2} & 35.4\pmstd{0.8} & 20.3\pmstd{0.6} & 77.2\pmstd{0.6} & 26.9\pmstd{1.1} & 37.0\pmstd{0.7} \\
\bottomrule
\end{tabular}
\caption{\textbf{Full component ablation table.} We show the full results and the mean and standard deviation for three seeds for Table 4.}
\label{tab:supp_ablation_full}
\end{table*}

{
    \small
    \bibliographystyle{ieeenat_fullname}
    \bibliography{main}
}

\end{document}